\documentclass[nonatbib]{article}
\usepackage{lipsum, lineno}
\usepackage{graphicx}
\usepackage[preprint]{neurips_2020}
\usepackage[utf8]{inputenc} % allow utf-8 input
\usepackage[T1]{fontenc}    % use 8-bit T1 fonts
\usepackage{hyperref}       % hyperlinks
\usepackage{amsmath}
\usepackage[noabbrev]{cleveref}
\usepackage{url}            % simple URL typesetting
\usepackage{booktabs}       % professional-quality tables
\usepackage{amsfonts}       % blackboard math symbols
\usepackage{nicefrac}       % compact symbols for 1/2, etc.
\usepackage{microtype}      % microtypography
\usepackage{lipsum, lineno}
\usepackage{setspace}
\usepackage{cite}
\usepackage[mathscr]{euscript}
\usepackage{placeins}
\usepackage{bm}
\usepackage{multirow}
\usepackage[table,xcdraw]{xcolor}
\usepackage[symbol]{footmisc}

\title{Shape modeling of longitudinal medical images: from diffeomorphic metric mapping to deep learning}

\author{
  Edwin Tay \\
  Delft University of Technology\\
  \texttt{e.w.s.tay@tudelft.nl} \\
  \AND
  Nazli Tümer\thanks{Both authors contributed equally to this study} \\
  Delft University of Technology\\
  \texttt{N.Tumer-1@tudelft.nl} \\
  \And
  Amir A. Zadpoor\footnotemark[1] \\
  Delft University of Technology\\
  \texttt{A.A.Zadpoor@tudelft.nl} \\
}

\begin{document}

\maketitle

\begin{abstract}

    Living biological tissue is a complex system, constantly growing and changing in response to external and internal stimuli. These processes lead to remarkable and intricate changes in shape. Modeling and understanding both natural and pathological (or abnormal) changes in the shape of anatomical structures is highly relevant, with applications in diagnostic, prognostic, and therapeutic healthcare. Nevertheless, modeling the longitudinal shape change of biological tissue is a non-trivial task due to its inherent nonlinear nature. In this review, we highlight several existing methodologies and tools for modeling longitudinal shape change (\textit{i.e.,} spatiotemporal shape modeling). These methods range from diffeomorphic metric mapping to deep-learning based approaches (\textit{e.g.,} autoencoders, generative networks, recurrent neural networks, \textit{etc.}). We discuss the synergistic combinations of existing technologies and potential directions for future research, underscoring key deficiencies in the current research landscape.
 
\end{abstract}

%\linenumbers

\section{Introduction}

	\textit{"Form follows function."}, although originally a perennial maxim coined by architect Louis Sullivan (1856-1924) in reference to pragmatic architectural design, has been adopted by the biomedical engineering community in reference to nature and its adaptability \cite{Sull96, RusMotAsh00}. This phrase is often used in reference to natural materials, which have optimized their shape and structures over millennia of evolution and adapted to their specialized tasks \cite{Wegst2014}. While studies have investigated both \textit{form} and its effect on \textit{function} \cite{Libonati2017, Wang2020mnmnmnmnm}, how it \textit{follows} remains nebulous. In particular, the way in which the shapes of anatomical structures change over time has long interested the biomedical engineering community, dating back to and even predating the seminal works of Darwin and Thompson \cite{Darwin2009, Thompson_1992}. Modeling and predicting the evolving characteristics of anatomical geometry is relevant, with applications for clinical diagnoses, prognoses, and interventional treatments. Therefore, uncovering the underlying processes governing shape change of anatomical structures over time remains a highly relevant and developing domain of research. 
			
\begin{figure}
		\centering
		\includegraphics[scale=0.9]{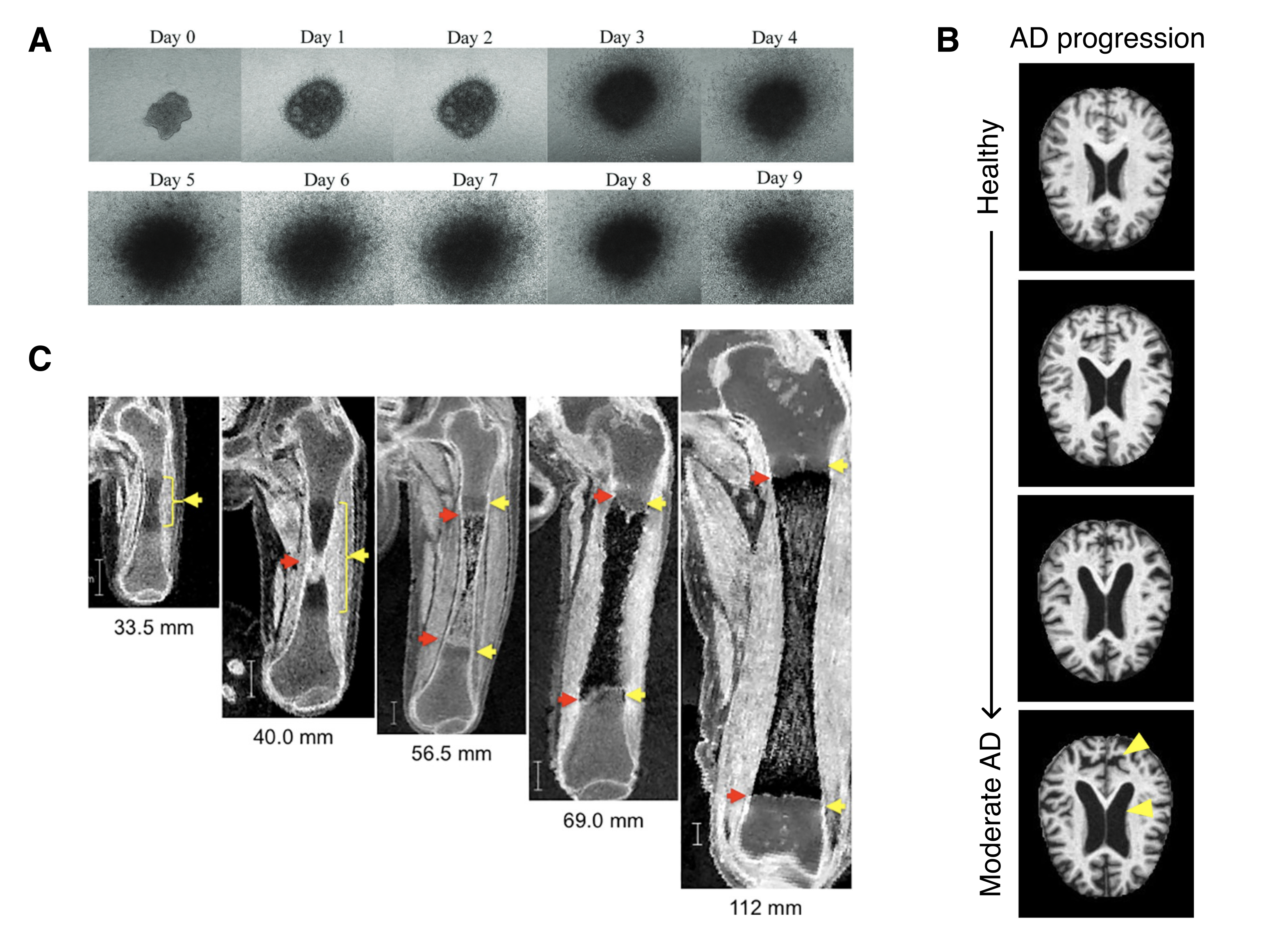}
		\caption{\textbf{A)} Longitudinal phase contrast imaging of 3D cell cultured cervical cancer spheroids \cite{Muniandy2021}*. \textbf{B)} Neurodegradation of brain structure with progression of AD, from healthy to moderate AD (top to bottom). Adapted from \cite{Pasnoori2024}*. \textbf{C)} Longitudinal MRI imaging of the morphogenesis of a femur during the embryonic and fetal periods. Figure adapted from \cite{Suzuki2019}*. *Images obtained from referenced sources and licensed under CC BY 4.0 (https://creativecommons.org/licenses/by/4.0/)}
		\label{fig:fig0}
	\end{figure}

	Longitudinal changes in the shapes of anatomical structures are relevant in a myriad of clinical applications, especially for early diagnosis and disease prognosis (Figure \ref{fig:fig0}). Developmental bone growth, for example, is a highly complex process wherein deficiencies or deviations from nominal standards could result in long-term health ramifications \cite{Parfitt2000, Weaver2014}. Some examples of such disorders include but are not limited to developmental hip dysplasia, osteogenesis imperfecta, scoliosis, and clubfoot \cite{Morcuende2003}. Early diagnosis could enable non-surgical treatments. Therefore, accurate ways of quantifying normal development and identifying abnormal variations is paramount \cite{Semler2019, Marzin2020, Newsome2016}. Another example is Alzheimer's disease (AD), one of the most common age-related neurodegenerative diseases \cite{Scheltens2021}. Commonly used techniques for early diagnosis of AD, such as neuropsychological tests, are unreliable and cerebrospinal-fluid biomarker measurements are intrusive and costly \cite{Alberdi2016}. In contrast, novel techniques examining structural brain changes from MRI can diagnose AD early and pre-symptomatically, while also informing future prognoses \cite{Mueller2005, Pegueroles2016, Blinkouskaya2021}. Yet another example is tumor growth, wherein growth rates and tumor sizes inform cancer severity and prognoses \cite{Clark1991, Morikawa2011, Kuroishi1990}. Thus, developing spatiotemporal growth models of tumors has been a long-standing field of research, ranging from early simplified deterministic 1-D models to more complex probabilistic simulations \cite{Adam1986, Jiang2005, Rejniak2010, Gerlee2013, Benzekry2014}. While not an exhaustive list, these examples demonstrate the wide-ranging applications and clinical relevance of developing robust spatiotemporal shape modeling tools and methodologies. 

	\FloatBarrier

	Early spatiotemporal shape modeling can be linked to morphometrics, wherein researchers attempted to analyze biological shape variation using statistical methods. Generally speaking, researchers analyzed variations of common anatomical landmarks across a population \cite{Slice2007}. These analyses examined variations in the coordinates of landmarks themselves, distances or relative angles between them, or metrics calculated from a combination thereof \cite{Rohlf90, Slice, 7e30a527-6aa0-3191-9009-c9d49643b4ef}. In the femur, for example, measurements such as the whole femur length, diaphyseal length, subtrochanteric anteroposterior and mediolateral diameters, anteroposterior physeal angles, alpha angle, and vertical diameter of the femoral head are some of the measurements used to characterize femoral anatomy \cite{Toogood2008, Wescott2005, Rissech2008}. These methods, however, are time-consuming and landmark placement can be unreliable, inconsistent, and fail to capture holistic spatial arrangement. Developments in computer vision and mathematical modeling tools have led to the development of computational anatomy (CA) \cite{Miller1997, Grenander1998, Miller2004}. Therein, the concept of shape manifolds and diffeomorphic transformations became central in describing anatomical shape variability over time. These tools have developed greatly in recent years. Furthermore, with the advent of deep learning (DL), novel methodologies have surfaced. Outlining available methodologies along with their strengths, weaknesses, and potential synergies is, thus, required. 

	In this review, we seek to highlight and discuss alternative techniques, methodologies, and tools used to model the changing shape of anatomical structures over time. For simplicity and due to the varied terminologies found in the literature, we use the terms spatiotemporal shape modelling and longitudinal shape models interchangeably. We also focus mainly on the techniques and tools themselves as opposed to their clinical applications. However, we highlight applications as necessary to enhance the descriptive value of the presented concepts. We also neglect exhaustive discussions on these methods' mathematical background and derivations, and instead refer the readers where necessary. Here, we refer to \textit{shape} in both a geometric sense (\textit{i.e.,} a set of points in $n$-dimensional Euclidean space ($\mathbb{R}^{n}$) with defined connections) and also in the intuitive sense of a visual boundary defining an object of interest within an image. This is important as both definitions play a role in the differing techniques we explore. A relatively similar review was carried out by Harie \textit{et al.}, however they explicitly focused on growth modeling and mainly discussed DL-based generative networks \cite{Harie2023}. In contrast, this review focuses on shape change over time in general, thus encompassing both growth and alternative biological processes (\textit{e.g.,} degeneration). Furthermore, this review does not focus exclusively on DL-based methods and also covers alternatives. This review begins with a discussion on diffeomorphisms and large deformation diffeomorphic metric mapping (LDDMM) framework, the most common early framework for spatiotemporal shape modeling. Then, we discuss deep learning-based tools, focusing on autoencoders, generative adversarial networks, recurrent neural networks, and transformers. Finally, we discuss the strengths and drawbacks of each tool generally, highlighting similarities and potential synergies. We also speculate on potential future outlooks and directions for research into spatiotemporal shape modelling. 
	
\section{Large Deformation Diffeomorphic Metric Mapping}
\label{sec:LDDMM}

	Of the many ways to describe variations in shapes in biology, a longstanding idea was first proposed by Thompson in his influential work \textit{"On Growth and Form"} in 1917 \cite{Thompson_1992}. Therein, he argued that variations in the shapes of biological organisms can be best described by geometrical transformations. This pioneering theory formed the basis for CA decades later with the development of computer vision and mathematical tools. In essence, CA assumes that individual shapes are described as diffeomorphic transformations of an underlying reference shape. As, in principle, an infinite number of diffeomorphisms can act on a reference shape, sets of diffeomorphisms can then be considered as an infinite dimensional manifold \cite{Marsland2020}. Accordingly, all the possible variations of a given shape can be represented within these manifolds, which are termed as 'shape spaces' \cite{Kendall1984, Monteiro2000, Rohlf2000}. These manifolds can then be enriched with Riemannian metrics which enable quantitative comparison of these shapes and further mathematical operations \cite{Younes2010, Miller2002, Miller2004}. This constitutes the basis for large deformation diffeomorphic metric mapping (LDDMM) framework \cite{Glauns2008, Durrleman2014}. Wherein, variations of anatomical shape in a population are described \textit{via} diffeomorphisms acting on an underlying reference template. These diffeomorphisms then make up the shape space, an infinite dimensional Riemannian manifold describing all possible variations of a shape in a population. The LDDMM framework can then be extended further for longitudinal shape modeling as we will discuss.  

		\subsection{Geodesics}
		
\begin{figure}[h!]
			\centering
			\includegraphics[scale=0.9]{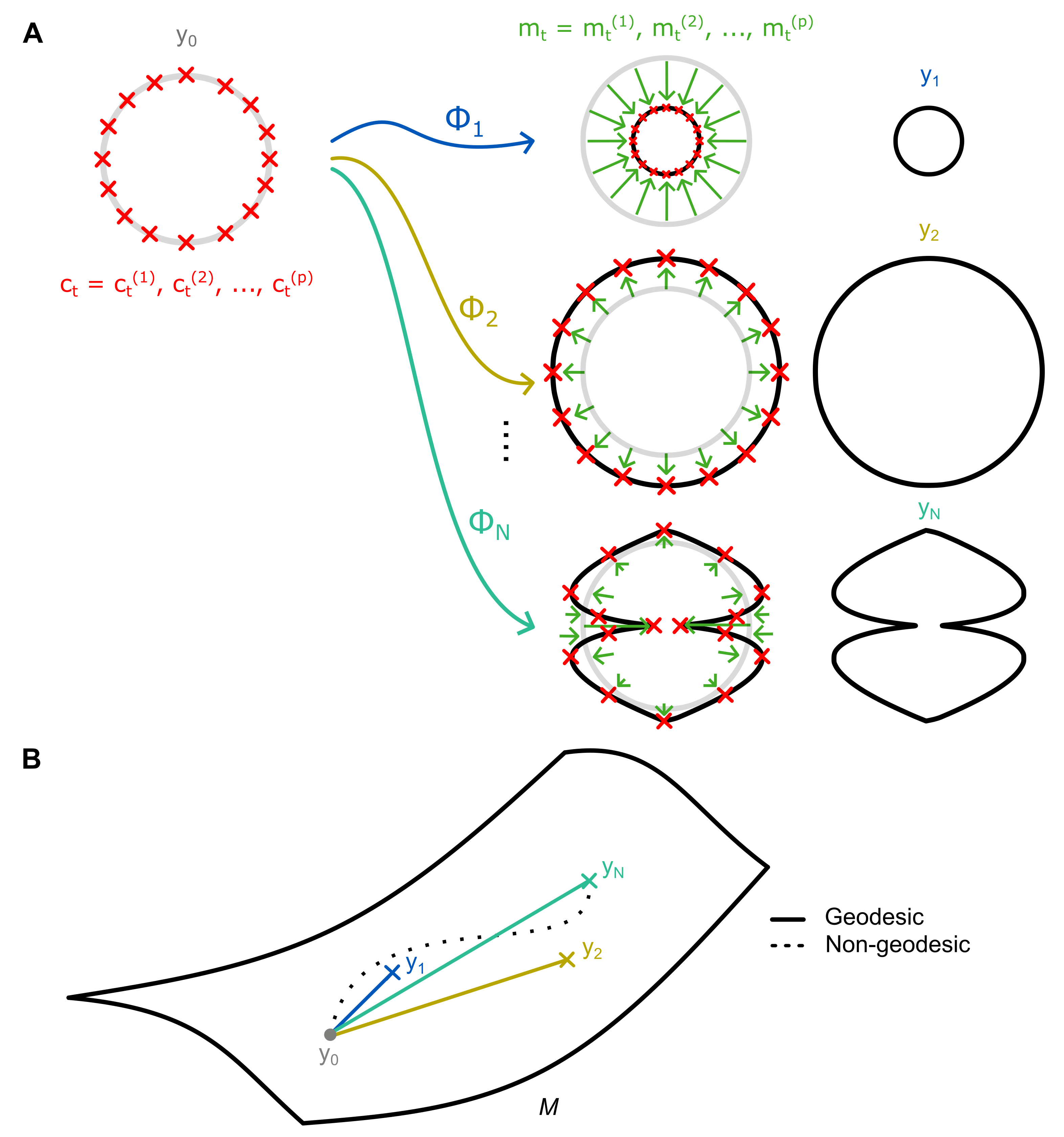}
			\caption{\textbf{A)} An illustration of diffeomorphisms $[\phi_{1},...,\phi_{N}]$ acting on a baseline reference shape $y_{0}$ to transform it to a shape within a dataset $[y_{1},...,y_{N}]$. A diffeomorphism constitutes $p$ momentum vectors $m_{t}$ acting on a similar number of control points $c_{t}$. \textbf{B)} Diffeomorphisms within the LDDMM framework lie on a Riemannian manifold $M$. The shortest paths (\textit{i.e.,} geodesics) connecting the reference shape and other shapes are used to describe the transformation and are determined based on minimal deformational energy.}
			\label{fig:fig1}
		\end{figure}

		For an initial reference shape $y_{0}$ and target shape $y_{1}$, a diffeomorphism $\phi_{1}$ exists which can be applied to transform the former to the latter (Figure \ref{fig:fig1}A). Following the convention of Bône \textit{et al.}, we denote this as $y_{1} = \phi_{1} \star y_{0}$ \cite{Bne2020}. Within the LDDMM framework, these diffeomorphisms follow the trajectories of a time-dependent vector field $t \rightarrow v_{t} \in C_{0}^{\infty} (\mathbb{R}^{d}, \mathbb{R}^{d})$ over [0,1]. Thus, starting from an identity mapping (Id), the change in $\phi_{t}$ over time ($\partial _{t} \phi_{t}$) is described as a functional composition of $v_{t}$ and $\phi_{t}$, denoted by $v_{t} \circ \phi_{t}$.(Equation \ref{eq:timevector}).
		
		\FloatBarrier
		\FloatBarrier
		
		\begin{equation}
			\partial _{t} \phi_{t} = v_{t} \circ \phi_{t} \quad \textrm{with} \quad \phi_{0} = \textrm{Id}.
			\label{eq:timevector}
		\end{equation}

		These diffeomorphisms can be difficult to obtain and describe, especially for complex shapes and deformations. Nevertheless, Miller \textit{et al.} demonstrated that these complex deformations could be succinctly described by utilizing the principle of conservation of momentum (\textit{i.e.,} vectors of momenta) \cite{Miller2006, Vaillant2004}. Specifically, $v_{t}$ can be discretized as a Gaussian convolution $g$ of $p$ momentum vectors $m_t = m^{(1)}_{t}, . . ., m^{(p)}_{t} \in \mathbb{R}^{d}$ acting over a set of corresponding control points $c_{t} = c^{(1)}_{t},..., c^{(p)}_{t} \in \mathbb{R}^{d}$, which also influence arbitrary points $x$ (Equation \ref{eq:vecconv}, Figure \ref{fig:fig1}A). 		
		
		\begin{equation}
			v_{t} : x \in \mathbb{R}^{d} \rightarrow \sum_{k=1}^{p} g[c_{t}^{(k)}, x] \cdot m_{t}^{(k)} \in \mathbb{R}^{d}
			\label{eq:vecconv}
		\end{equation}
		
		In general notation, the Gaussian kernel function is defined as $g: x, x' \in \mathbb{R}^{d} \rightarrow \exp \vert\vert x' - x \vert\vert_{l^2}^{2} / \sigma^2$, with kernel width $\sigma > 0$. Solutions for $\phi_{t}$ are non-unique due to the infinite-dimensional nature of the underlying shape space manifold. Thus, the geodesic, that is the diffeomorphism requiring the least amount of deformational energy (Equation \ref{eq:defenergy}, Figure \ref{fig:fig1}B), is utilized \cite{Miller2002, Durrleman2014}. 
		
		\begin{equation}
			\frac{1}{2} \int_{t=0}^{1} \vert\vert v_{t} \vert\vert_{G_{c_{t}}}^{2} = \frac{1}{2} \int_{t=0}^{1} m_{t}^{T} \cdot G_{c_{t}} \cdot m_{t}
			\label{eq:defenergy}
		\end{equation}
		
		, where, $G_{c_{t}}$ is the $p \times p $ kernel symmetric positive-definite matrix of general term $g[c_{t}^{(k)}, c_{t}^{(l)}]$ and $(\cdot)^{T}$ denotes a matrix transposition. These geodesics' control points and momenta are also fully determined by their initial values and the following Hamiltonian equations (Equation \ref{eq:hamiltonian}). The former observation is particularly notable as, then, the system of initial momenta and control point locations $S_{0} = \{ c_{0}, m_{0}\}$, fully parametrize the entire flow of diffeomorphisms. 
		
		\begin{equation}
			\dot{c_{t}} = G_{c_{t}} \cdot m_{t}  \quad ; \quad \dot{m_{t}} = -\frac{1}{2} \nabla_{c_{t}} \{  m_{t}^{T} \cdot G_{c_{t}} \cdot m_{t}   \} 
			\label{eq:hamiltonian}
		\end{equation}	
		
		, where $\nabla_{c_{t}}$ is the gradient operator with respect to $c_{t}$. Thus, a Riemannian manifold $\mathscr{D}_{c_{0}}$, based on Equations \ref{eq:timevector}, \ref{eq:vecconv}, and \ref{eq:hamiltonian}, can be described as follows (Equation \ref{eq:riemanmanifold}): 
				
		\begin{equation}
			\begin{aligned}
			\mathscr{D}_{c_{t}} = \{ \phi_{1} \vert &\partial_{t}\phi_{t} = v_{t} \circ \phi_{t}, \quad \phi_{0} = \mathrm{Id}, \quad v_{t} = \mathrm{Conv}(c_{t}, m_{t}), \\
			&(\dot{c_{t}}, \dot{m_{t}}) = \mathrm{Ham} (c_{t}, m_{t}), \quad m_{0} \in \mathbb{R}^{p \times d}  \}
			\end{aligned}
			\label{eq:riemanmanifold}
		\end{equation}

		\subsection{Geodesic Regression}
			
			In representing each shape in a longitudinal dataset as a diffeomorphism of a template shape, the challenge remains in establishing the relationships between each shape. This is particularly important as deriving any underlying relationships between different shapes and independent variables (\textit{i.e.,} time) is essential for spatiotemporal shape modeling. Acquiring these relationships \textit{via} standard regression techniques, for instance, is non-trivial due to the non-Euclidean structure of the Riemannian manifolds of diffeomorphisms. Nonetheless, Fletcher proposed an extension of standard linear regression to be applicable in a manifold-based setting, termed geodesic regression \cite{fletcher2011, ThomasFletcher2012}. Their technique was then developed further for a variety of applications, but the developments of Fishbaugh \textit{et al.} for use in longitudinal shape modelling are of particular interest \cite{Fishbaugh2013, Fishbaugh2017} for our purposes. 
			
\begin{figure}[h!]
				\centering
				\includegraphics[scale=0.9]{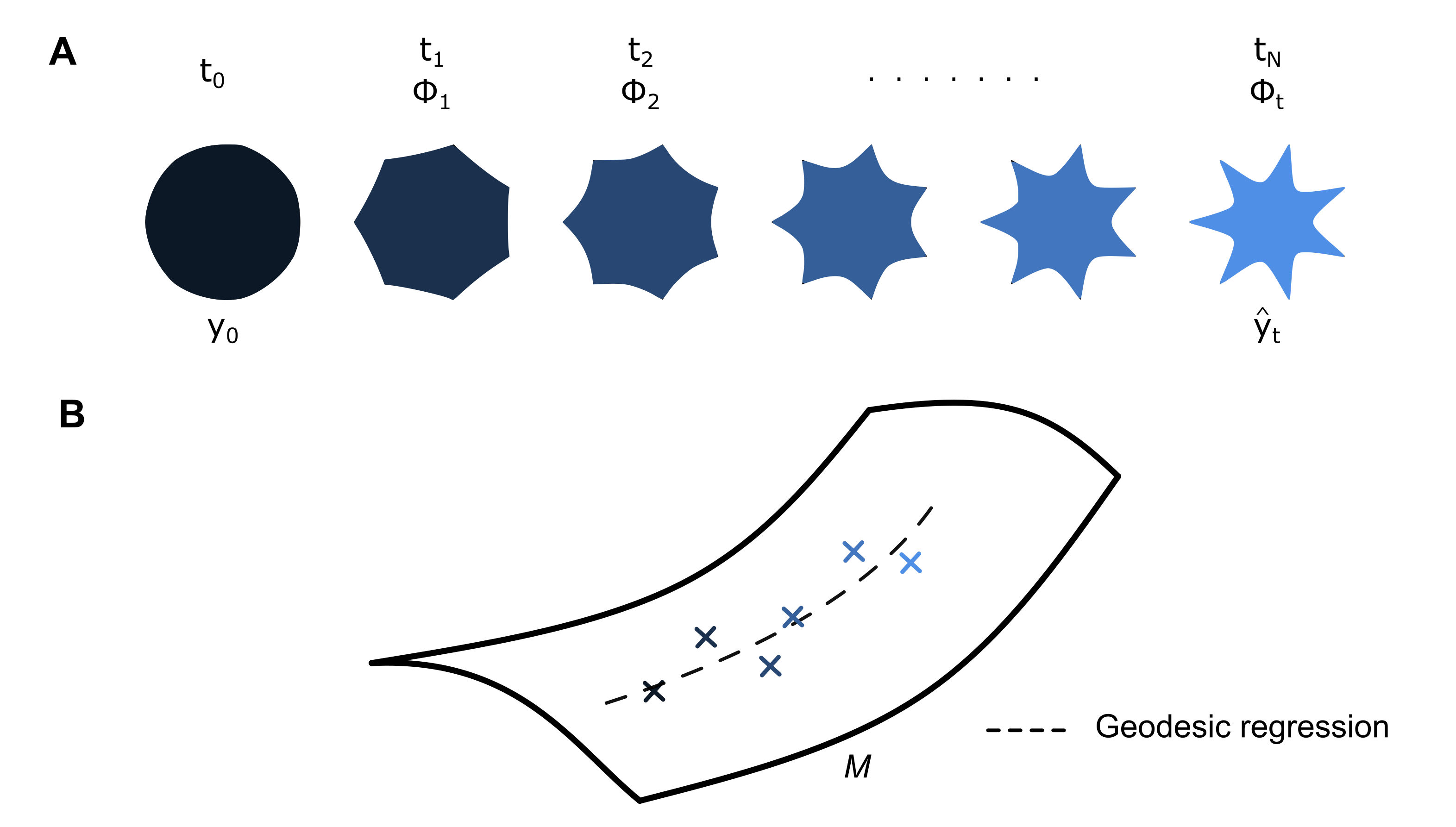}
				\caption{\textbf{A)} Each shape in a longitudinal dataset of $N$ shapes spanning $[t_{0},t_{N}]$ can be described with a corresponding diffeomorphism at time $t$, $\phi_{t}$, acting on reference shape $y_{0}$. These diffeomorphisms are obtained from the estimation of an underlying group-average geodesic. Thus, the action of $\phi_{t}$ on $y_{0}$ leads to an estimate for the corresponding shape $\hat{y_{t}}$ \textbf{B)} Each diffeomorphism lies on a Riemannian manifold $M$, and an underlying group-average geodesic, which describes the trajectory of diffeomorphisms, can be estimated \textit{via} geodesic regression.}
				\label{fig:fig2}
			\end{figure}

			In detail, for a longitudinal dataset of shapes with $N$ number of observations in the time range $[t_{0}, t_{N}]$, shape change over time is taken as a baseline shape $y_{0}$ being continuously deformed at each time point $t$ by a corresponding diffeomorphism $\phi_t$ (Figure \ref{fig:fig2}A). In principle, $\phi_t$ should lead to the baseline shape morphing to completely match the observed shape $y_{t} = \phi_{t} \star y_{0}$. However, in this context of estimating a holistic group-average geodesic (Figure \ref{fig:fig2}B), we note that a diffeomorphism instead leads to an estimation of the observed shape at time $t$ instead $(\hat{y_{t}} = \phi_{t} \star y_{0})$. A regression criterion can then be expressed as follows (Equation \ref{eq:shapregression}), subject to Equation \ref{eq:vecconv} and \ref{eq:hamiltonian} \cite{Fishbaugh2013, Fishbaugh2017}. 
			
			\begin{equation}
				E(y_{0}, S_{0}) = \sum_{i=1}^{N_{}} \frac{1}{2 \gamma^{2}} \vert\vert (\phi_{t_{i}} \star y_{0}) - y_{t_{i}} \vert\vert^{2} + L(S_{0})
				\label{eq:shapregression} 
			\end{equation}
			
			, where, $L(S_{0})$ represents a regularity term for the time-varying deformation, determined by the kinetic energy of the control points at $S_{0}$ (Equation \ref{eq:defenergy}). $\gamma^2$ represents a term used to balance the importance between the data and regularity terms. Thus, given a dataset of longitudinal shapes, during the minimization of Equation \ref{eq:shapregression} the baseline shape $y_{0}$, initial control point locations $c_{0}$, and initial momenta $m_{0}$ are the parameters which are estimated. This general form of geodesic shape regression was then developed further to incorporate both shape and image data based on a weighted joint optimization routine \cite{Fishbaugh2014}. Their multimodal approach demonstrated improved performances as opposed to exclusive shape or image approaches. Nevertheless, optimization schemes to solve for the underlying geodesic regressions are computationally expensive, especially for large-scale image datasets. Recently, Ding \textit{et al.} have also proposed methods to enhance their speed and effectiveness using DL \cite{Ding2019}. They demonstrated that the use of encoder-decoder networks with GPU acceleration could increase computation speeds, enabling the scaling up of studies towards larger datasets encompassing more subjects or longer timescales. Developments notwithstanding, these methods were limited to single subject regressions; geodesic regression up to this point has mainly captured spatiotemporal variability of only single subjects, thus these methods were extended further to capture populations and their intervariabilities. 

			\subsection{Hierarchical Models} 
			
\begin{figure}[h!]
					\includegraphics[scale=0.9]{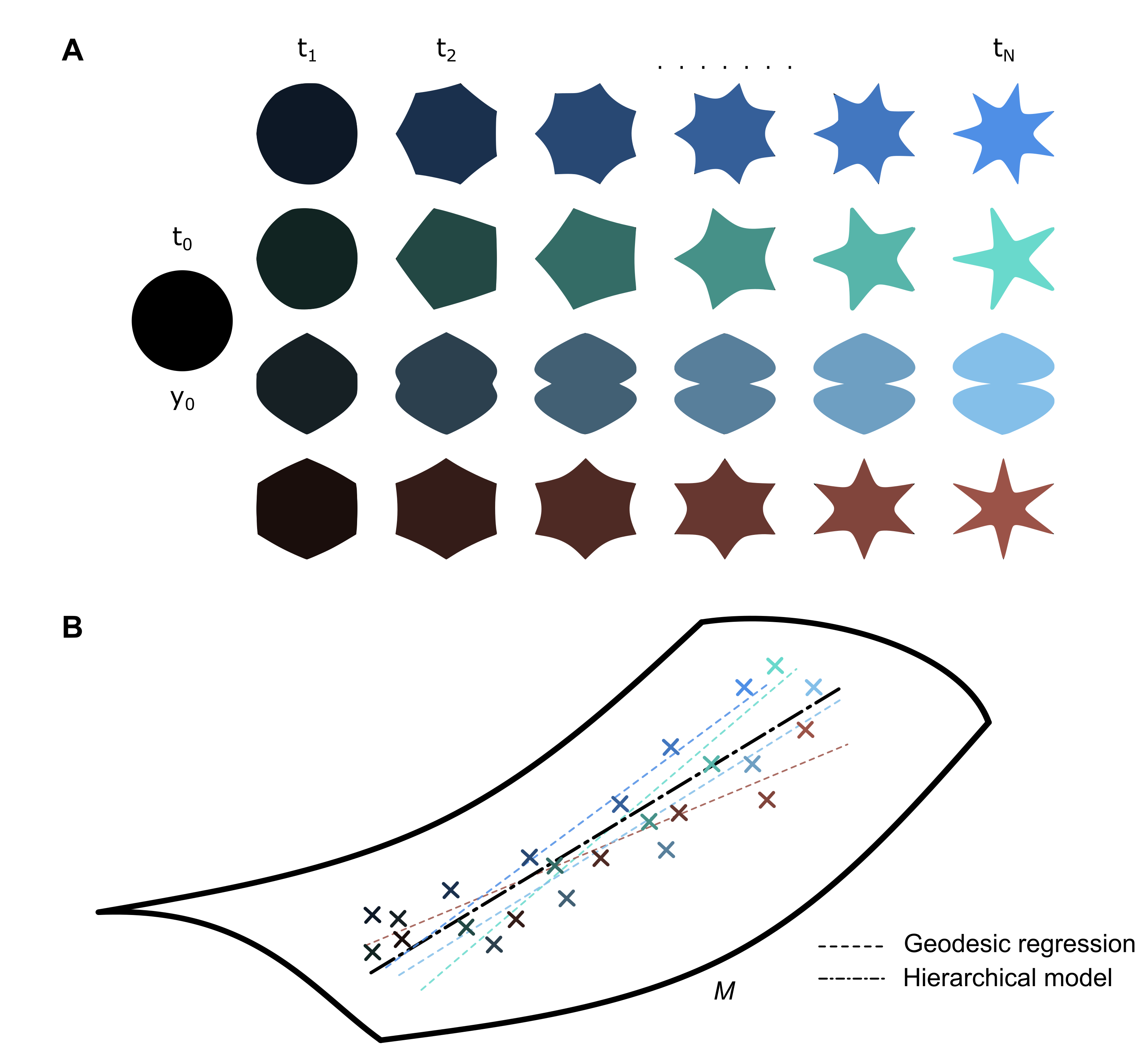}
					\caption{\textbf{A)} A dataset of various shapes spanning $[t_{0},t_{N}]$ can be described as diffeomorphic transformations of an underlying baseline template shape $y_{0}$. \textbf{B)} Individual shape trajectories can be modeled by individual geodesic regressions, which can be used to estimate a group-average geodesic or vice-versa (\textit{i.e.}, group-average geodesic used to estimate individual trajectories). }
					\label{fig:fig3}
				\end{figure}

				While geodesic regression can describe an object's longitudinal trajectory over time, it is insufficient to describe the longitudinal characteristics of multiple objects in a large dataset (Figure \ref{fig:fig3}A). Thus, the LDDMM framework was further extended towards hierarchical models. Early work done by Muralidharan \textit{et al.} could estimate an underlying groupwise mean geodesic based on individual geodesics (Figure \ref{fig:fig3}B) \cite{Muralidharan2012}. They did this with a least squares estimation of the underlying mean geodesic, using Sasaki metrics to compare individual trends. This was developed further by Singh \textit{et al.} as a generalization of hierarchical linear models to a manifold-based setting \cite{Singh2015}. Schiratti \textit{et al.} took a slightly different modeling approach, wherein they first found the underlying group-average spatiotemporal trajectory and represented individual trajectories within the dataset as space and time transformations of this group-average \cite{SchirattiAllassonni2017}. This approach offers more flexibility as, unlike the former approach, it is not heavily dependent on initial time point choice, easing time reparametrization. Bône \textit{et al.} developed this approach further for shape data within the LDDMM framework specifically \cite{Bone2018, Bne2020}.

				Briefly, the hierarchical generative longitudinal models of Bône \textit{et al.} rely on exp-parallelization ($\text{ExpP}_{\gamma}^{v_{i}}$) and a time warp function ($\psi_{i}$) to account for individual spatial and temporal differences respectively (Equation \ref{eq:hierarchicalmodel}) \cite{Bone2018, Bne2020}. Exp-parallelization essentially offers a tool to define parallel curves on a manifold whilst retaining the underlying structure \cite{Schiratti2015, SchirattiAllassonni2017}. This enables us to define individual trajectories traversing the manifold as a variation of a group average. A time warp, on the other hand, accounts for the temporal characteristics of each individual's trajectory (\textit{i.e.}, onset time, and rate of progression).
				
%				\begin{equation}
					\begin{align}
						\text{ExpP}_{\gamma}^{v_{i}} &[\psi_{i}(t_{i,j})] \star y_{0} \mathop{\sim}\limits^{\text{iid}} \mathcal{N}_{\epsilon}(y_{i,j}, \sigma_{\epsilon}^{2})\label{eq:hierarchicalmodel} \\
						\text{where} \quad & \vert\psi_{i} : t \rightarrow \alpha_{i} \cdot (t- \tau_{i}) + t_{0} \nonumber \\ 
						& \vert v_{i} = \mathrm{Conv}(c_{0}, m_{i}), \quad m_{i} = A_{0, m_{0}^{\perp}} \cdot s_{i} \nonumber
                     \end{align}
%				\end{equation}
				
				In turn, each component of the model (Equation \ref{eq:hierarchicalmodel}) is as follows. A prediction for shape observation $j$ of subject $i$, $y_{i,j}$ is modeled as a noisy estimate with variance $\sigma_{\epsilon}^2$. $y_{i,j}$ itself is predicted as a diffeomorphic transformation of a baseline reference shape $y_{0}$ transformed by an underlying group average geodesic $\gamma$ space-shifted by exp-parallelization to match an individual's trajectory $v_{i}$. The time warp function $\psi_{i}$ accounts for temporal characteristics, where $\alpha_{i}$ denotes progression rate, $\tau_{i}$ is onset time, and $t_{0}$ is the reference time. $v_{i}$ accounts for the individuals' spatial variability and, in essence, is Equation 2 with some additional constraints. Namely, the momenta $m_{i}$ are obtained from a mixing matrix $A_{0, m_{0}^{\perp}}$ and $q$ source parameters $s_{i} = s_{i}^{(1)}, ..., s_{i}^{q}$. The parameters to be estimated which define individual trajectories are modeled as independent samples from normal distributions: 
				
%				\begin{equation}
					\begin{align}
						\alpha_{i} &\mathop{\sim}\limits^{\text{iid}} \mathcal{N}_{[0, +\infty]} (1, \sigma_{\alpha}^2) \\
						\tau_{i} &\mathop{\sim}\limits^{\text{iid}} \mathcal{N}(t_{0}, \sigma_{\tau}^2) \nonumber \\
						s_{i} &\mathop{\sim}\limits^{\text{iid}} \mathcal{N}(0, 1) \nonumber	
						\end{align}
%				\end{equation} 

				Taken together, a mixed effects model can be defined for the gathered parameters.
				Fixed effects, which account for parameters affecting the trajectories of all the subjects, can be denoted as $\theta = (\theta_{1}, \theta_{2})$. Where, $\theta_{1} = (t_{0}, \sigma_{\tau}, \sigma_{\alpha}, \sigma_{\epsilon})$ and $\theta_{2} = (y_{0}, c_{0}, m_{0}, A_{0})$. The random effects $z_{i}$ account for variations for each subject, where $z_{i} = (\alpha_{i}, \tau_{i}, s_{i})$. This nonlinear multi-parameter optimization task is computationally complex and expensive and relies on a multi-step \textit{calibration}, \textit{personalization}, and \textit{simulation} scheme detailed further in Bône \textit{et al.} \cite{Bne2020}. In brief, it utilizes a novel Monte Carlo Markov Chains-Stochastic Approximation Expectation Maximization-Gradient Descent (MCMC-SAEM-GD) algorithm detailed further in the reference. 

				\subsection{Applications and Further Works}
				
					Overall, the use of hierarchical models provides us with a structured framework to characterize longitudinal data, both on an individual and group level. The use of a group-average trajectory enables us to quantify the variation of an individual's progression from a normative scenario \cite{Kim2017}. This also has the potential for prognostic benefits. For example, Cury \textit{et al.} could detect shape changes in the thalamus of patients suffering from dementia 10 years prior to clinical symptoms by comparing healthy and diseased spatiotemporal trajectories \cite{Cury2016}. Bône \textit{et al.} have demonstrated the use of exp-parallelization and time reparametrization to transport a population average trajectory onto new subjects \cite{Bne2017}. Thus, they demonstrated that population-average normative trajectories can be leveraged to predict trends in shape change or disease progression for new, unseen subjects. Similarly, Koval \textit{et al.} implemented a manifold-based hierarchical model but in the context of graph networks \cite{Koval2018}. Specifically, they derived a population-based estimate for cortical atrophy dynamics and demonstrated the capability to characterize patient-specific atrophy dynamics. They further extended this work to account for multimodal data such as biomarker levels and cognitive impairment scores to develop a comprehensive spatiotemporal atlas of Alzheimer's disease \cite{Koval2021}. This method of integrating the use of biomarkers (\textit{i.e.,} genetic and clinical factors) alongside imaging has gained traction and not only demonstrates soundness in and of itself \cite{Dalca2015} but also has the potential to enhance the predictive capacity of existing frameworks with multimodality. Couronne \textit{et al.} further demonstrated the efficacy of multimodal models in the context of Parkinson's disease prognosis \cite{Couronne2019}. Utilizing both imaging and neurophysiological test score data, they demonstrated the robustness and efficacy of multimodality to improve predictive performance.
					
					Nevertheless, the hierarchical model framework is still being developed further to refine its modeling efficacy and integrate newer technological developments. As opposed to modeling correlations along a manifold as quasi-linear in the manner of geodesics, Hanik \textit{et al.} proposed to utilize generalized Bézier curves to model nonlinear relationships with the rationale that many biological processes are nonlinear (\textit{e.g.,} cardiac motion) \cite{Hanik2022}. Their initial work demonstrated the potential for extending this principle further and potentially decomposing longitudinal trends (\textit{i.e.,} disease progression) into different components of a nonlinear curve, enabling more granular analyses. Hong \textit{et al.} also investigated the effects of subject-specific characteristics by including multivariate intercept models in their formulation of a hierarchical geodesic model \cite{Hong2019}. Debavelaere \textit{et al.} developed a methodology to investigate datasets with heterogeneous populations (\textit{i.e.,} a dataset with diverging longitudinal dynamics) \cite{Debavelaere2020}. They developed an unsupervised algorithm that is able to detect clusters of subgroups within a dataset and differentiate their trajectories, accounting for diverging or converging trajectories from a population normal. Furthermore, the advent of DL has led to augmentations of the LDDMM framework due to its increased computational efficiency of processing large datasets\cite{Yang2023, BenAmor2022}. Bône \textit{et al.} demonstrated the use of autoencoders to learn an atlas and class of diffeomorphisms that describe a dataset of shapes and meshes \cite{Bne2019}. They further extended their work to also account for the texture (\textit{i.e.,} appearance) of images \cite{Bne2020a}. Pathan and Hong also demonstrated the potential of using DL to learn vector momenta utilized in the LDDMM framework \cite{pathan2018}. Other novel developments include the utilization of implicit neural representations (INRs) \cite{NEURIPS2020_53c04118}. Dummer \textit{et al.} demonstrated the potential of using INRs to extend the LDDMM framework towards increased robustness and resolution independence \cite{dummer2023}. 
				
					To surmise, the LDDMM framework is a powerful tool for representing and modeling a dataset of shapes. Assuming an underlying template shape, the LDDMM framework represents individual shapes as diffeomorphic transformations of this template. These diffeomorphisms lie on an infinite dimensional Riemannian manifold, thus relying on geodesic regression and parallel transport tools to estimate the longitudinal trajectories traversing the underlying data manifold. Hierarchical models can then be utilized to model differing spatiotemporal trajectories of a population, capable of estimating population average spatiotemporal trajectories and also quantifying intra and inter-individual differences. Whilst, in recent years, the proliferation of DL-based techniques has seemingly eclipsed LDDMM-based techniques, the framework is continuously developing. In fact, many of the developments seek to utilize DL tools to accelerate the framework and increase its efficacy. LDDMM methods are readily available in several software packages and applications such as Deformetrica \cite{Bne2018}, Leaspy (https://leaspy.readthedocs.io/en/stable/), and Morphomatics \cite{Morphomatics}.  
					
\section{Deep Learning}
\label{sec:DL}

	In medical imaging, DL-based solutions have pushed the state-of-the-art further for a variety of tasks. From image segmentation, disease diagnosis, and prognosis to synthetic image synthesis, DL represents a powerful paradigm for the future of medical image analysis \cite{Shen2017, Wang2020dfdfdfdfd}. In this section, we highlight alternative network architectures that have been utilized for spatiotemporal shape modeling.
		
	\subsection{Autoencoders}
	
\begin{figure}
		\centering
		\includegraphics[scale=0.9]{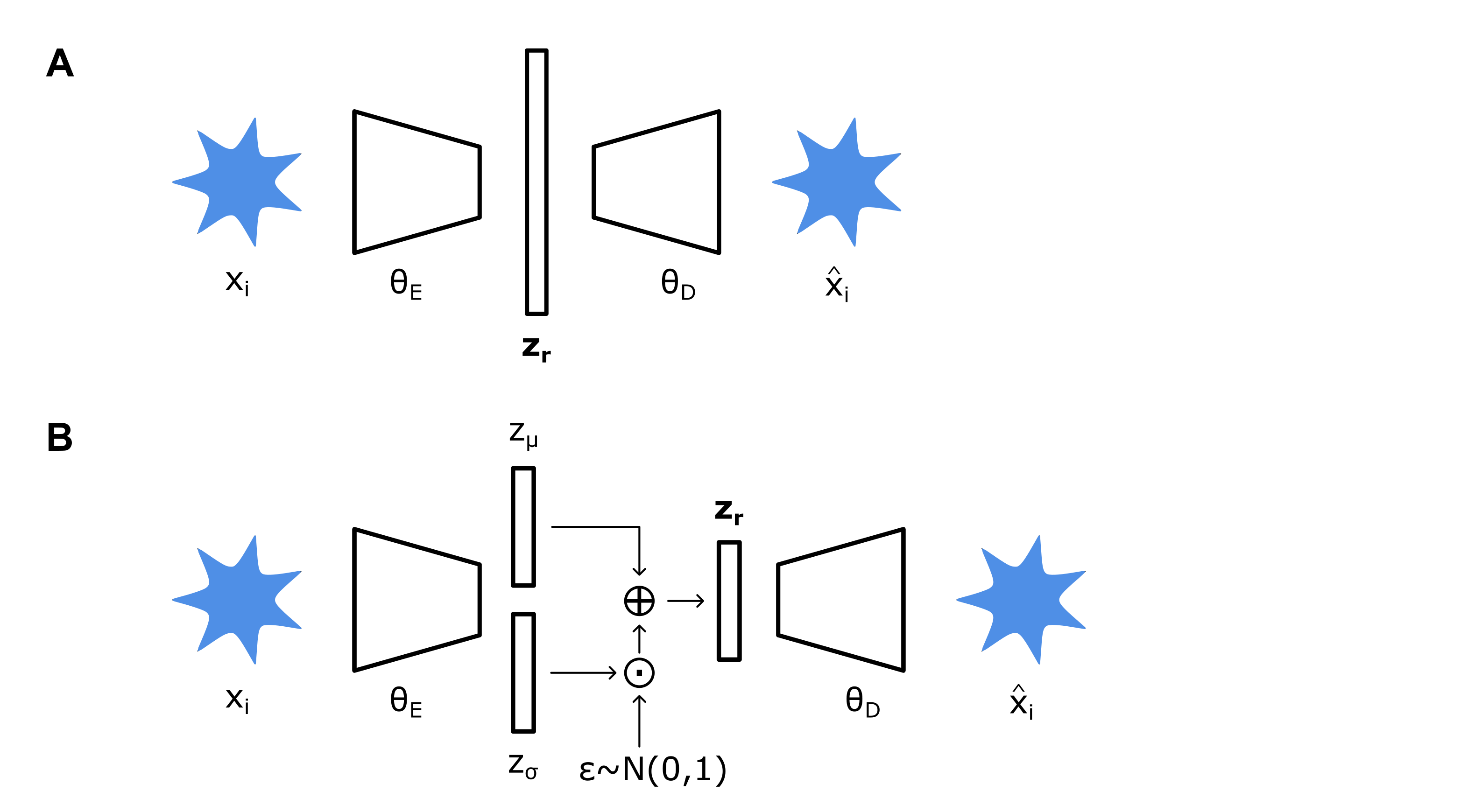}
		\caption{\textbf{A)} Autoencoder structure consisting of an encoder ($\theta_{E}$) which translates an input image $x_{i}$ into a vector of latent variables $\mathbf{z_{r}}$. A decoder ($\theta_{D}$) then attempts to reconstruct input data $\hat{x}_{i}$ from $\mathbf{z_{r}}$. \textbf{B)} A variational autoencoder consists of similar components, however $\theta_{E}$ maps $x_{i}$ instead to deterministic parameters $z_{\mu}$ and $z_{\sigma}$ which describe a probabilistic distribution. These are then used to obtain $\mathbf{z_{r}}$ and similarly decoded.}
		\label{fig:fig4}
	\end{figure}

	Autoencoders (AEs) are a neural network (NN) architecture consisting of an encoder and decoder module (Figure \ref{fig:fig4}A). This architecture, in principle, seeks to compress data to a low-dimensional latent space, reducing them to $r$ number of latent variables, $\bm{z_{r}}$. These latent variables themselves can then be utilized for other tasks as they represent, in essence, a compressed low-dimensional representation of higher-dimensional data. Thus, the weights of the encoder $\theta_{E}$ and decoder $\theta_{D}$ modules are learned to accurately de-construct input data down into a latent representation and re-construct them into the original input data, respectively \cite{LopezPinaya2020}. The objective when training an AE is then to minimize the loss function $\mathcal{L}^{rec}$, which takes the form of a dissimilarity function or reconstruction loss, to find $\theta_{E}$ and $\theta_{D}$ (Equation \ref{eq:autoencoder}). A commonly used loss function is the L2 norm  (Equation \ref{eq:l2norm}), but alternatives exist that could work better for different forms of input data \cite{Khare2022}.   
	
%	\begin{equation}
		\begin{align}
			\min \bm{\mathcal{L}}(\theta_{E}, \theta_{D}) &= \min_{\theta_{E}, \theta_{D}} \sum_{i=1}^{N} \mathcal{L}^{rec}(x_{i}, \hat{x_{i}}) \label{eq:autoencoder} \\ 
			\mathrm{where} \quad \hat{x_{i}} &= \theta_{D}(\theta_{E}(x_{i})) \nonumber
        \end{align}
%	\end{equation}
	
	\begin{equation}
		\mathcal{L}^{rec}(x, \hat{x}) = \vert\vert x - \hat{x} \vert\vert_{2}^{2}
		\label{eq:l2norm}	
	\end{equation}

	Latent variables $\bm{z_{r}}$ in a standard AE configuration are, in principle, unstructured. Thus, sampling new data in generative processes might not lead to valid data as the properties and ranges of $\bm{z_{r}}$ values are not known. Appending regularization terms, $\mathcal{L}^{reg}(\bm{z_{r}})$, to the cost function (Equation \ref{eq:autoencoder}) can lead to more structured latent variables and the spaces they inhabit (Equation \ref{eq:regautoencoder}). These modified AE structures are, thus, referred to as regularized autoencoders \cite{Ehrhardt2022}. 
	
%	\begin{equation}
		\begin{align}
			\min \bm{\mathcal{L}}(\theta_{E}, \theta_{D}) &= \min_{\theta_{E}, \theta_{D}} \sum_{i=1}^{N} \mathcal{L}^{rec}(x_{i}, \hat{x_{i}}) + \lambda \mathcal{L}^{reg}(\bm{z_{i}}) \label{eq:regautoencoder} \\ 
			\mathrm{where} \quad \hat{x_{i}} &= \theta_{D}(\theta_{E}(x_{i})) \nonumber
        \end{align}
%	\end{equation}
	
	$\lambda$ is a balancing term used to adjust the trade-off between latent-space regularity and reconstruction quality. On the other hand, $\mathcal{L}^{reg}(\bm{z_{r}})$, can take many forms and is detailed further elsewhere \cite{Ehrhardt2022, pmlr-v48-arpita16}. 
	
	A variation of AEs is variational autoencoders (VAEs) which are similar but treat encoding and decoding in a probabilistic manner (Figure \ref{fig:fig4}B)\cite{RezendeMohamed14}. Instead of directly mapping input data to latent variables, VAEs map input data to probabilistic distributions of their corresponding latent variables. Briefly, $\theta_{E}$ maps input data to deterministic parameters, mean $z_{\mu}(x)$ and standard deviation $z_{\sigma}(x)$, which describe an underlying probabilistic distribution (usually Gaussian) of the latent space. These deterministic parameters are then injected with stochasticity sampled from a fixed normal distribution, where $\odot$ denotes a Hadamard product (Equation \ref{eq:vae}). This configuration is necessary to preserve the stochasticity within the latent space while enabling gradient-based backpropagation during training \cite{Ehrhardt2022}. In turn, the loss function (Equation \ref{eq:autoencoder}) is now modified to consider both reconstruction quality and regularity of the latent space (Equation \ref{eq:vaeloss}). The latter is usually represented by a Kullback–Leibler divergence, detailed elsewhere \cite{Ehrhardt2022}. Overall, a probabilistic treatment of latent variables and the spaces they inhabit leads to more structured, compact, and continuous latent spaces. This, in turn, leads to a smoother sampling of latent variables for generative processes and representation learning in general. 
	
	\begin{equation}
		z = z_{\mu}(x) + z_{\sigma}(x) \odot \epsilon \quad   \mathrm{with} \ \epsilon \sim \mathcal{N}(0,1)	
		\label{eq:vae}
	\end{equation}
	
	\begin{equation}
		\mathcal{L}^{VAE} = \mathcal{L}^{rec} + \mathcal{L}^{KL}
		\label{eq:vaeloss}
	\end{equation}
		
	AEs, VAEs, and variations thereof have many applications in generative frameworks and tasks involving reduced dimension representations of high dimensional data such as images. The strengths of these architectures in modeling complex data within low-dimensional representations could lend themselves well to capturing the complex nonlinearities inherent in longitudinal datasets. Mouches \textit{et al.} utilizes an invertible latent space disentanglement module within an autoencoder framework to determine latent variables that affect age-related changes \cite{pmlr-v143-mouches21a}. Isolated age-related latent variables can then be varied, with age-unrelated components kept constant, to simulate the aging of a particular individual. Following a similar vein, Zhao \textit{et al.} utilized a cosine-based loss function to disentangle brain age from image representation \cite{Zhao2021}. They did so with a self-supervised learning methodology, optimizing the correspondence between the 'directionality' of latent variables in the latent space and physical developmental trajectories. Sauty and Durrleman utilized a VAE to learn latent variables representing images within a longitudinal dataset \cite{Sauty2022}. These latent variables are then fitted to a linear longitudinal mixed-effects progression model similar to those of the LDDMM framework.  
	
	Overall, AEs and VAEs represent powerful tools for reducing high-dimensional data into a low-dimensional latent space, efficiently encapsulating longitudinal data into compressed latent variables. Nevertheless, the problem remains with the underlying physical significance of latent variables and latent spaces, or the lack thereof \cite{Ehrhardt2022}. Latent variables and their spaces are solely reduced dimension representations of the original input data; latent variables have no underlying physical meaning. For example, the distribution of latent variables has been shown to be affected by training parameters, demonstrating their capricious nature \cite{Lapenda2020}. Thus, latent variables cannot be considered spatiotemporal variables. However, rational structuring and regularization to ensure that latent spaces are enriched with physical meaning can lead to better outcomes. Another point of concern is that both AEs and VAEs generally treat latent spaces and variables in a Euclidean manner, when in fact research has shown that a manifold-based approaches may be more prudent \cite{pmlr-v130-connor21a}. These problems remain active fields of research, with solutions such as regularization and explicitly structuring latent spaces deterministically being continually developed \cite{yeyeyeyeye, ghosh2020from}. Nonetheless, existing works for spatiotemporal shape modeling demonstrated the potential applicability for autoencoders and learned latent variables to model longitudinal trajectories.	
	
	\subsection{Generative Adversarial Networks}
				
\begin{figure}
			\centering
			\includegraphics[scale=0.9]{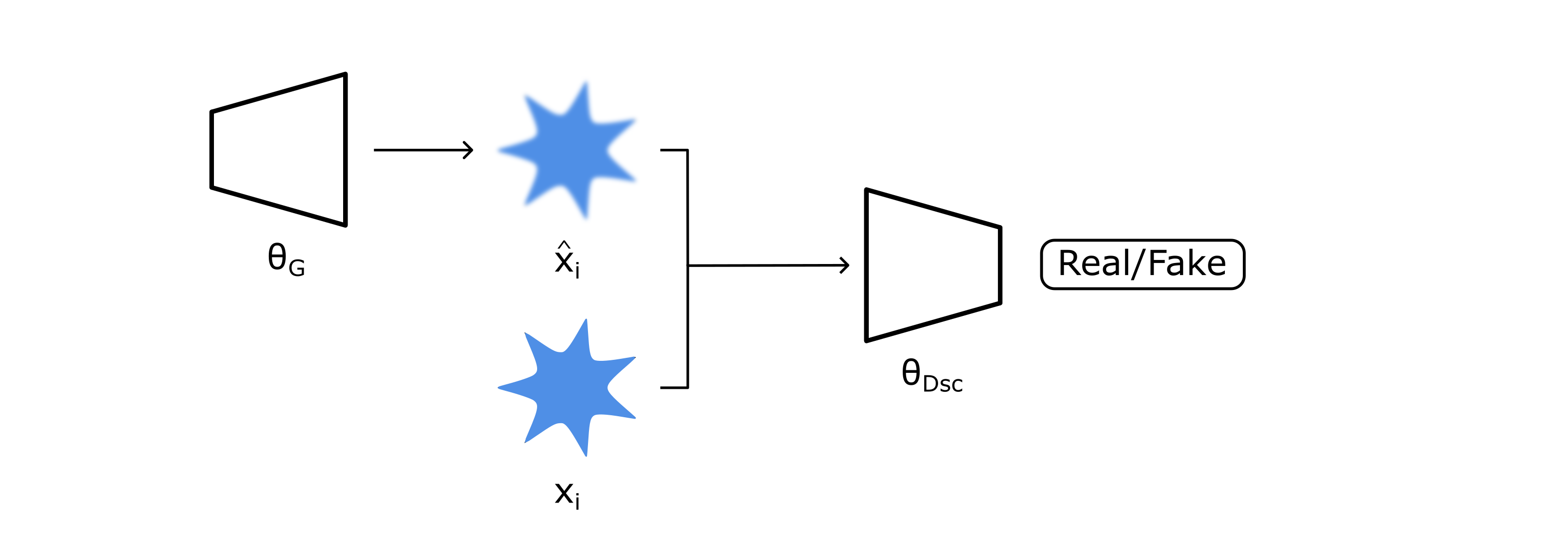}
			\caption{A generative adversarial network (GAN) architecture consists of a generator ($\theta_{G}$) which creates synthetic data ($\hat{x}_{i}$) resembling real data ($x_{i}$). A discriminator ($\theta_{Dsc}$) then attempts to differentiate real versus 'fake' synthetic data. Both $\theta_{G}$ and $\theta_{Dsc}$ are jointly trained so that the former generates increasingly realistic images while the latter is able to discriminate real \textit{vs.} fake data better.}
			\label{fig:fig5}
		\end{figure}

		Generative Adversarial Networks (GANs) are neural network architectures first proposed by Goodfellow \textit{et al.} \cite{NIPS2014_5ca3e9b1}. In principle, they consist of generator $\theta_{G}$ and discriminator $\theta_{Dsc}$ networks being trained simultaneously (Figure \ref{fig:fig5}). Therein, the former is trained to create new synthetic images, whilst the latter is trained to detect if an image is real or fake. In detail, $\theta_{G}$ maps random input variables $\nu$ (sampled from a prior distribution $p(\nu)$) to the data space in an attempt to generate data $\hat{x}_{G}$ resembling data from a real dataset $x_{r}$. In turn, both types of data are fed into $\theta_{Dsc}$, whereby $\theta_{Dsc}$ is trained to determine if data is real or fake. The objective function used to train both networks simultaneously is then a minimax problem (Equation \ref{eq:GANminimax}).
				
		\begin{equation}
			\min_{\theta_{G}}\max_{\theta_{Dsc}} V (\theta_{G}, \theta_{Dsc}) = E_{x\sim p_{data}(x_{r})} [ \log D(x) ] + E_{z\sim p(\nu)} [ \log (1- D(G(\nu)))]
			\label{eq:GANminimax}
		\end{equation} 
		
		This architecture is very powerful, as the adversarial training configuration leads to the generator module being capable of generating realistic synthetic images that are indistinguishable from real data \cite{Wang2017, Gui2023}. Trained generators are then useful for many applications. In the context of medical imaging, examples include image synthesis, segmentation, and classification, among others \cite{Yi2019}. Many variants exist, and more are continually being developed, for which the reader is referred to other papers for further details \cite{Jabbar2021}. In the context of spatiotemporal shape modeling, GANs represent a powerful tool. Similar to previously discussed AEs and VAEs, their generative capacity can potentially be utilized to capture the underlying spatiotemporal trajectories. 
			
		Elazab \textit{et al.} used a stack of 3D GANs to predict brain tumor growth \cite{NL2016150352}. Specifically, with an input image and physiological feature maps, a generator predicted a brain scan at the proceeding time point whose accuracy was evaluated by a discriminator. Their results outperformed contemporary methods but relied on stacking and training consecutive GANs, which is computationally inefficient. Alternatively, Zhang \textit{et al.} utilized GANs to uncover the underlying data manifold of longitudinal progression for face aging \cite{zhang2017age}. They first encode images to latent vectors, which are concatenated with age-related feature vectors and then mapped onto a manifold. Discriminators ensure regularized latent vector generation and image realism of the generators. Based on this, Ravi \textit{et al.} developed a 2D framework to model age-related brain degeneration in the context of Alzheimer's diagnosis \cite{RaviAlex2019}. They incorporated further voxel-based and region-level constraints which acted as biological constraints to model Alzheimer's progression, leading to improved prognoses. They developed this work further to examine 3D MRIs for a more holistic view of the brain \cite{Ravi2022}. Utilizing a 3D training consistency mechanism and a super-resolution module led to a full 4D model of brain aging without a loss in anatomical detail. Following the same principle of temporal embedding within a latent space, Schön \textit{et al.} similarly implemented a GAN-based network for embedding temporal directionality in generators \cite{SchonSelvan23}. Alternative GAN architectures have also been investigated. Wasserstein GANs (WGANs) utilize Wasserstein distances as a loss function as opposed to regularly used Jensen-Shannon divergence \cite{pmlr-v70-arjovsky17a}. This architecture leads to more stable training outcomes and was utilized by Wegmayr \textit{et al.} as a recursive generator model to predict time steps in brain aging \cite{Wegmayr2019}. Combined with a classifier network, they present a framework for both predicting aged brain images and Alzheimer's prognosis, outperforming standard methods. In StyleGAN and derivatives thereof, the principle of style transfer and additional, intermediate latent spaces is utilized to improve generator architectures and disentangle latent space components and their effects on synthesized images \cite{Karras_2019_CVPR, Karras_2020_CVPR, Fetty2020}. Han \textit{et al.} developed a framework for image-based osteoarthritis prognosis using StyleGAN as the generative architecture \cite{Han2022}. This enabled them to construct the underlying manifold of longitudinal knee aging, and furthermore, they demonstrated that their model outperforms human radiologists in early diagnosis of osteoarthritis. Similarly, Gadewar \textit{et al.} utilized StarGAN-v2, a similar style-based generator architecture, to predict aging in structural MRIs of the brain \cite{Gadewar2023, Gadewar2023a}. 

		In short, GAN-based architectures and adversarial training represent powerful tools for spatiotemporal shape modeling. In particular, discriminators support the structuring and regularization processes so that the latent space of generator modules is physically meaningful, similar to previously discussed regularized AEs. While GANs have their own challenges in terms of training stability, mode collapse, convergence, and image fidelity, continual developments in training schemes, architectures, and loss functions have led to continuous improvements \cite{Yi2019, Gui2023, Saxena2021}. Generators with well-defined and structured latent spaces and rational generative processes enable us to predict growth trajectories. Said structuring of latent spaces is facilitated by discriminators and loss functions, which allow us to ensure smooth latent spaces that are temporally consistent and valid. In essence, in helping structure latent spaces, discriminators implicitly define the underlying manifold of spatiotemporal shape progression. Similar to previously discussed AEs, this structuring process ensures that latent spaces and variables therein can be endowed with meaningful physical characteristics.   			

	\subsection{Recurrent Neural Networks}
	
\begin{figure}
			\centering
			\includegraphics[scale=0.9]{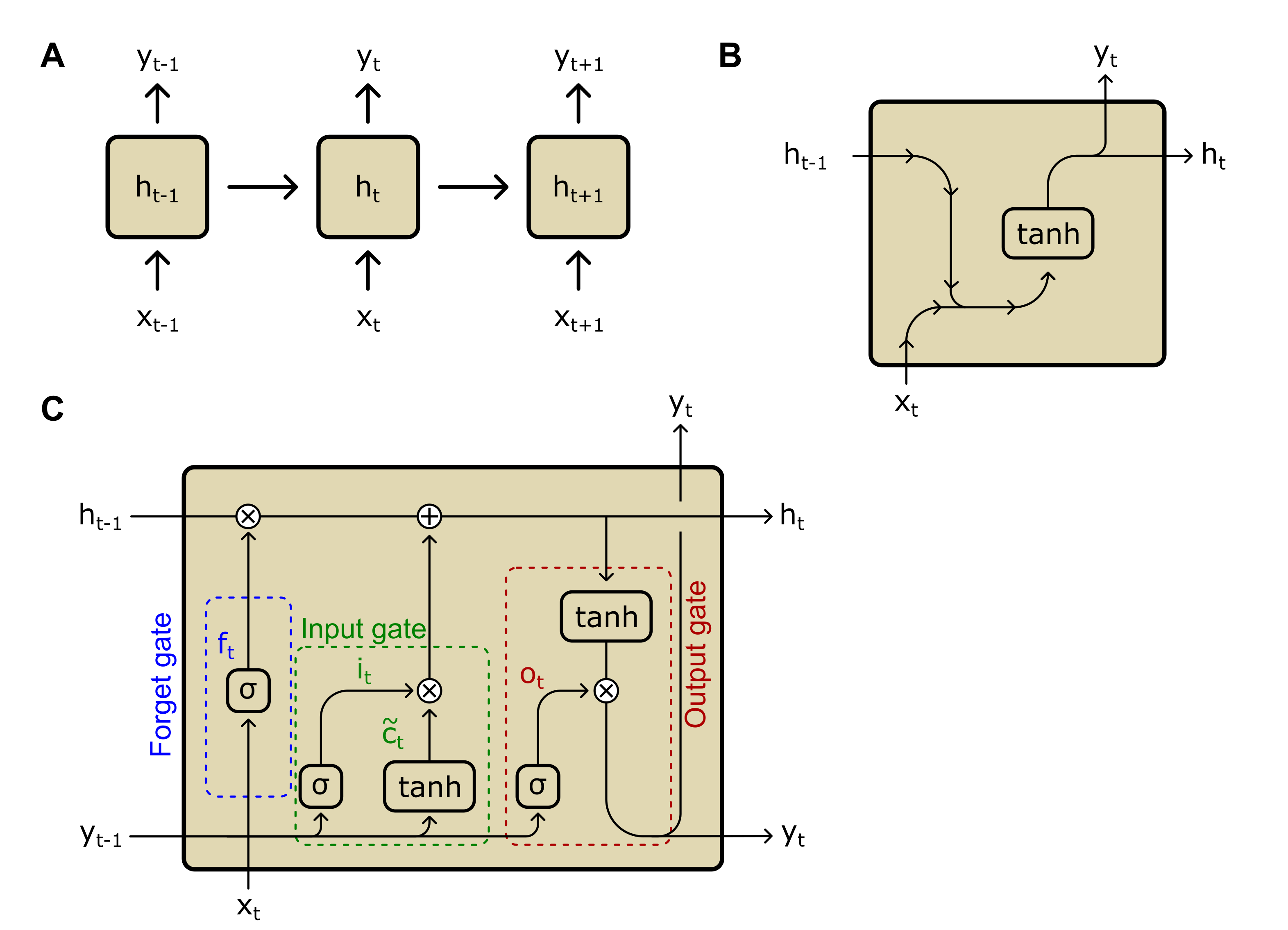}
			\caption{\textbf{A)} A recurrent neural network (RNN) is trained along a sequence of time points $t$. Based on input ($x_{t}$) and output data ($y_{t}$), a hidden state ($h_{t}$) is continuously updated using context units. \textbf{B)} A simple context unit in early RNNs, wherein the input from the previous time point of the hidden state ($h_{t-1}$) is combined with $x_{t}$ and a $\tanh$ activation function to calculate $y_{t}$ and update $h_{t}$. \textbf{C)} A Long Short-Term Memory (LSTM) cell is a more complex unit used to maintain $h_{t}$ and determine $y_{t}$. Both sigmoidal ($\sigma$) and $\tanh$ activation functions are used within forget, input, and output gates, which determine which data is removed, added, and output from the cell. Figure inspired by existing work of \cite{45500}.}
			\label{fig:fig6}
		\end{figure}

		Recurrent neural networks (RNNs) are a type of NN that are used to model sequential data such as a time series \cite{Lipton2015, Salehinejad2018, Staudemeyer2019}. They do so by considering data along a whole sequence's trajectory during training and inference; RNNs are designed explicitly with features that connect and consider data inputs across longitudinal sequences by maintaining memory (\textit{i.e.,} a hidden internal state $h_{t}$ which is continually updated at each time point $t$). Early RNNs utilized simple 'context units', which are units independently connected to nodes in the hidden layer of an NN (Figure \ref{fig:fig6}B) \cite{Elman1990}. These context units are then updated along steps in a data sequence \textit{via} activation functions as the RNN is trained along a sequence (Figure \ref{fig:fig6}A). To address practical issues regarding optimization of network parameters (exploding/vanishing gradients during backpropagation) \cite{Lipton2015, Salehinejad2018, Staudemeyer2019}, newer RNN architectures utilize complex memory cells instead of traditional nodes \cite{Hochreiter1997, Yu2019, Chung2014}. A prime example is the Long Short-Term Memory (LSTM) cell. These types of cells utilize 'gates', components that determine when and how to modify the memory of the cell (Figure \ref{fig:fig6}C). In detail, the three main gates of a standard LSTM cell are the forget, input, and output gates. The forget gate utilizes the output of the previous time point $y_{t-1}$ and the current input $x_{t}$ with a sigmoidal activation function to determine data in $h_{t}$ should be 'forgotten' (Equation \ref{eq:forgetgate}). Where $W$ and $b$ denote the weights and biases of the threshold units, respectively. 

		\FloatBarrier

		\begin{equation}
			f_{t} = \sigma(W_{f}[y_{t-1}, x_{t}] + b_{f})
			\label{eq:forgetgate}
		\end{equation} 
		
		Similarly, for the input gate, a sigmoidal activation function determines which values are to be updated (Equation \ref{eq:lstmit}), and a $\tanh$ activation function creates new values to be added to the internal state (Equation \ref{eq:lstmtanh}).   
		
		\begin{equation}
			i_{t} = \sigma(W_{i}[y_{t-1}, x_{t}] + b_{i})
			\label{eq:lstmit}
		\end{equation}
		
		\begin{equation}
			\tilde{c_{t}} = \tanh (W_{c}[y_{t-1}, x_{t}] + b_{c})
			\label{eq:lstmtanh}
		\end{equation}
			
		Together, Equations \ref{eq:forgetgate} - \ref{eq:lstmtanh} are combined to update $h_{t}$, where $*$ denotes pointwise multiplication (Equation \ref{eq:lstminternalupdate}).
		
		\begin{equation}
			h_{t} = f_{t} * h_{t-1} + i_{t} * \tilde{c_{t}}
			\label{eq:lstminternalupdate}
		\end{equation} 
			
		Finally, the output is determined by the output gates using $x_{t}$ and $y_{t-1}$ (Equation \ref{eq:lstmoutput1}) and $h_{t}$ (Equation \ref{eq:lstmoutput2}). 
		
		\begin{equation}
			o_{t} = \sigma ( W_{o}[y_{t-1}, x_{t}] + b_{o})
			\label{eq:lstmoutput1}
		\end{equation}
		
		\begin{equation}
			y_{t} = o_{t} * \tanh (h_{t})
			\label{eq:lstmoutput2}
		\end{equation}
			
		Thus far, LSTMs have been used for natural language processing or other tasks examining relatively low-dimensional data. In the context of images and CNNs, LSTMs have been adapted for image inputs in the form of the convolutional LSTM (ConvLSTM) \cite{NIPS2015_07563a3f}. ConvLSTM is able to capture temporal information and dependencies in a sequence of images while ensuring that spatial information is preserved during encoding. This has led to its use and marked effectiveness in video prediction tasks \cite{Lotter2016, Lu2017}. 
		
		In the context of longitudinal medical imaging, RNNs have improved the outcomes of segmentation \cite{Gao2018} and disease stage classification tasks \cite{Santeramo2018, Gao2018a, Cui2019, Ouyang2021, Ding2023}. In explicitly modeling shape change using ConvLSTMs and its derivatives, however, RNNs have seen comparatively less uptake potentially due to the significantly high GPU memory requirements \cite{MaZhangLiu22}. Some studies nevertheless utilize RNNs as components within larger frameworks to avoid this obstacle. For example, Pathan and Hong used LSTMs to predict the vector momentum sequences to deform a longitudinal baseline image in an LDDMM framework \cite{SharminYi18}. This approach leverages the effectiveness of the LDDMM framework to predict changes over time without loss of detail and the computational efficiency of DL. Louis \textit{et al.} utilized RNNs to encode longitudinal trajectories into a latent space \cite{Louis2019}. These encoded trajectories are then decoded to construct the manifold and the Riemannian metrics lying on this manifold. Ma \textit{et al.} utilized ConvLSTMs alongside a transformer in a 'growth prediction module' to predict tumor growth \cite{Ma2023}. They demonstrated that utilizing both components in a unified module leads to better-predicted growth morphologies. Zhang \textit{et al.} extended the ConvLSTM framework with the goal of modeling spatiotemporal sequences (ST-ConvLSTM) \cite{Zhang2020}. Their ST-ConvLSTM units learn both temporal and spatial dependencies in a sequence; for a 3D image slice, ST-ConvLSTM learns both the changes over time for that slice and accounts for the adjacent slices. 
		
		To surmise, RNNs represent a powerful network architecture for capturing temporal dependencies within a longitudinal dataset. Nevertheless, the issue of high GPU memory requirements for imaging data persists. This particular requirement precludes the use of RNNs for longitudinal shape modeling. Nevertheless, Ma \textit{et al.} sought to address this by developing multi-scale RNN frameworks, which demonstrably improve performance with much lower GPU memory costs \cite{MaZhangLiu22, Ma2023a}. Chen \textit{et al.} demonstrated the use of signed distance function-based representations with ConvLSTMs to predict longitudinal changes in the shape of vestibular schwannoma \cite{ChenWolterJelmerNeve2024}. They demonstrated a proof of concept for using signed distance functions, which could address issues of large memory requirements of conventional ConvLSTMs operating directly on images. All in all, developments in using LSTMs for medical imaging datasets are relatively recent and have yet to be fully investigated in the context of longitudinal medical image shape modeling. 
	
	\subsection{Transformers}
		
\begin{figure}
			\centering
			\includegraphics[scale=0.9]{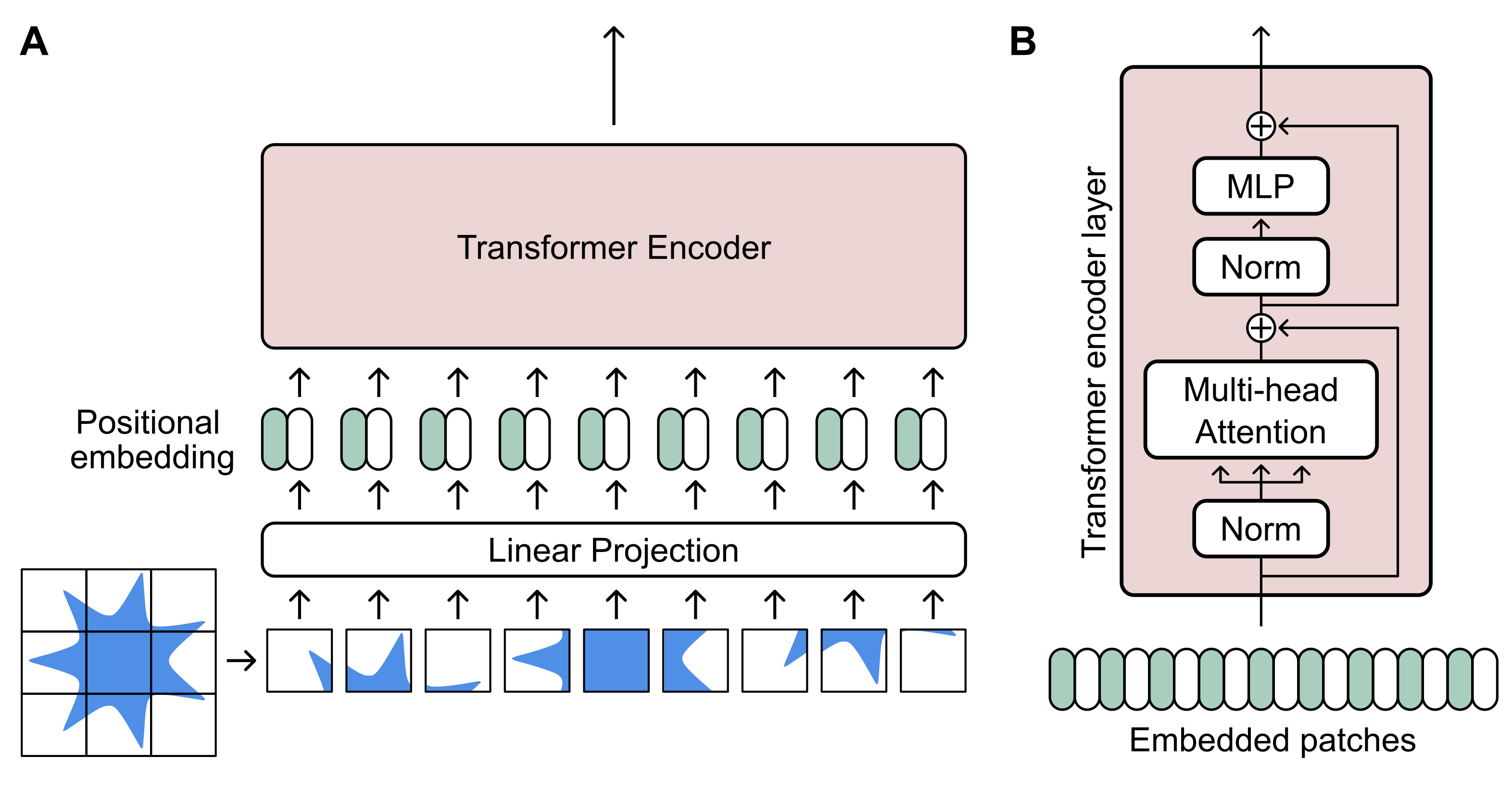}
			\caption{\textbf{A)} A Visual Transformer (ViT) architecture tokenizes an input image by first delineating it into smaller patches. Each patch is then linearly projected and embedded alongside its positional data before being fed into a transformer encoder. \textbf{B)} A transformed encoder layer takes the embedded image patches as tokens and uses them within a multi-head attention-based encoder layer. Figure inspired by existing work of \cite{Dosovitskiy2020}.}
			\label{fig:fig7}
		\end{figure}

		Transformers are a relatively recent development in DL. Originally designed for natural language processing (NLP) tasks \cite{Vaswani2017}, they utilize a novel attention mechanism based on saliency, which can capture long-range dependencies in data sequences. The architecture was later developed further specifically for image data with the Vision Transformer (ViT) framework \cite{Dosovitskiy2020}. In any case, transformer networks rely firstly on tokenization of input data \cite{Shamshad2023, Islam2024, Torralba2024-ce}. For text data, tokenization is relatively straightforward, but for images, the ViT framework relies on patching, a process to divide a larger image into $N$ smaller fixed-sized patches. These image patches are then linearly projected and embedded alongside their positions to encode the relative positions of each patch. These embedded patch and position data represent and thus are token representations of the original image data $\mathbf{T} = [t_{i},..., t_{N}]$ (Figure \ref{fig:fig7}). These tokens are then passed to attention models,  where 'query' $\mathbf{q}$, 'key' $\mathbf{k}$, and 'value' $\mathbf{v}$ vectors for each token are calculated by their respective projection matrices ($W_{q}, W_{k}, W_{v}$) (Equation \ref{eq:qkv}). These vectors are then compiled into larger matrices $\mathbf{Q}, \mathbf{K}, \mathbf{V}$ respectively (Equation \ref{eq:qkv}). 
		
		\FloatBarrier

%		\begin{equation}
			\begin{align}
				\mathbf{Q} &= 
				\begin{bmatrix}
					\mathbf{q}_{1}^{\top}\\
					\vdots \\ 
					\mathbf{q}_{N}^{\top}
				\end{bmatrix} = 
				\begin{bmatrix}
					(W_{q}t_{1})^{\top}\\
					\vdots \\ 
					(W_{q}t_{N})^{\top}
				\end{bmatrix} = \mathbf{T} \mathbf{W}_{q}^{\top} \nonumber \\ 
				\mathbf{K} &= 
				\begin{bmatrix}
					\mathbf{k}_{1}^{\top}\\
					\vdots \\ 
					\mathbf{k}_{N}^{\top}
				\end{bmatrix} = 
				\begin{bmatrix}
					(W_{k}t_{1})^{\top}\\
					\vdots \\ 
					(W_{k}t_{N})^{\top}
				\end{bmatrix} = \mathbf{T} \mathbf{W}_{k}^{\top} \label{eq:qkv} \\ 			
				\mathbf{V} &= 
				\begin{bmatrix}
					\mathbf{v}_{1}^{\top}\\
					\vdots \\ 
					\mathbf{v}_{N}^{\top}
				\end{bmatrix} = 
				\begin{bmatrix}
					(W_{v}t_{1})^{\top}\\
					\vdots \\ 
					(W_{v}t_{N})^{\top}
				\end{bmatrix} = \mathbf{T} \mathbf{W}_{v}^{\top}\nonumber 
			\end{align}
%  		\end{equation}

		These vectors are then used to calculate an attention score which represents the saliency of the tokens $\mathbf{T}_{out}$ in an architecture referred to as self-attention (Equation \ref{eq:selfattention}), where $\sqrt m $ is a scaling factor based on the dimensionality of $\mathbf{K}$ to ensure stability during training.
		
		\begin{equation}
			\mathbf{T}_{out} = \mathrm{softmax} \left(\frac{\mathbf{Q}\mathbf{K}^{\top}}{\sqrt m} \right) \mathbf{V}
			\label{eq:selfattention}
		\end{equation}
		
		However, self-attention is relatively limited and, instead, multi-head self-attention (MHSA) is able to better capture map saliency and capture context. Essentially, MHSA has $n$ parallel self-attention layers, each with their own learned projection matrices $\mathbf{W}_{q,i}, \mathbf{W}_{k,i}, \mathbf{W}_{v,i}$. Outputs from each layer are concatenated and projected using a learned output projection matrix $\mathbf{W}_{o}$ (Equation \ref{eq:multihead}). 
		
%		\begin{equation}
			\begin{align}
				&\mathbf{T}_{out,multi}	= \mathrm{concat}(\mathbf{T}_{out,1}, ..., \mathbf{T}_{out,n}) \mathbf{W}_{o} \label{eq:multihead}\\
				\mathrm{where} \quad &\mathbf{T}_{out,i} = \mathrm{softmax} \left(\frac{\mathbf{Q}_{i}\mathbf{K}_{i}^{\top}}{\sqrt m} \right) \mathbf{V}_{i} \nonumber \\ 
				\mathrm{and} \quad 	\mathbf{Q}_{i} &= 
				\begin{bmatrix}
					\mathbf{q}_{1,i}^{\top}\\
					\vdots \\ 
					\mathbf{q}_{N,i}^{\top}
				\end{bmatrix} = 
				\begin{bmatrix}
					(W_{q,i} \; t_{1})^{\top}\\
					\vdots \\ 
					(W_{q, i} \; t_{N})^{\top}
				\end{bmatrix} = \mathbf{T} \mathbf{W}_{q,i}^{\top} \nonumber \\ 
				\mathbf{K}_{i} &= 
				\begin{bmatrix}
					\mathbf{k}_{1,i}^{\top}\\
					\vdots \\ 
					\mathbf{k}_{N,i}^{\top}
				\end{bmatrix} = 
				\begin{bmatrix}
					(W_{k,i} \; t_{1})^{\top}\\
					\vdots \\ 
					(W_{k,i} \; t_{N})^{\top}
				\end{bmatrix} = \mathbf{T} \mathbf{W}_{k,i}^{\top} \nonumber \\ 			
				\mathbf{V}_{i} &= 
				\begin{bmatrix}
					\mathbf{v}_{1, i}^{\top}\\
					\vdots \\ 
					\mathbf{v}_{N, i}^{\top}
				\end{bmatrix} = 
				\begin{bmatrix}
					(W_{v,i} \; t_{1})^{\top}\\
					\vdots \\ 
					(W_{v,i} \; t_{N})^{\top}
				\end{bmatrix} = \mathbf{T} \mathbf{W}_{v,i}^{\top} \nonumber
            \end{align}
%		\end{equation}		
		
		These tokenized representations and attention modules are then integrated into various NN architectures and can be configured for many applications, especially in medical image analysis \cite{Azad2024}. In particular, the capability to capture long-range dependencies and focus on salient features across long input sequences could potentially be applicable for predicting shape changes over time sequences. 
		
		In the context of longitudinal shape modeling, the use of transformers are still relatively unexplored. Sarasua \textit{et al.} was one of the first to apply transformers to model longitudinal shape trajectories \cite{Sarasua2021}. They forecasted the change in the shape of meshes of the left hippocampus in an encoder-decoder-style architecture utilizing a bidirectional transformer encoder. They extended this work further by explicitly embedding Alzheimer's cognitive impairment scores and utilizing pre-trained transformers \cite{Sarasua2022}. The latter method revolved around freezing most layers of a pre-trained transformer and fine-tuning it on a selected task to decrease the number of trainable parameters \cite{Lu2021}. The former method of embedding cognitive scores was also similarly utilized by Xia \textit{et al.} to synthesize longitudinal brain images \cite{Xia2021}. With an input baseline brain image, their transformer architecture embeds a health state and age progression to synthesize changes over time. To improve the quality of their predicted progressions, they trained their networks in an adversarial manner with additional loss functions to preserve subject identity. Wang \textit{et al.} developed a comprehensive transformer-based framework to predict tumor growth \cite{Wang2022}. Their so-called static-dynamic framework utilizes a transformer-based module to first encode and enhance high-level features of detected tumors. Then, a transformer-based growth estimation module is employed to predict growth based on the aforementioned extracted features. 
		
		Nonetheless, the transformer architecture is still in its relative infancy, with applications for predicting longitudinal shape trajectories still developing. Advances in transformer architectures, such as incorporating multi-scale convolutions for enhanced time-series prediction, could potentially be applied to imaging data as well \cite{Wang2023}. However, there are a number of caveats to the enhanced performance of transformer-based networks \cite{Li2023}. Firstly, the nature of the transformer architecture leads to lower degrees of inductive bias, necessitating larger amounts of training data for better performance. This could potentially be addressed with pre-training as demonstrated by Lu \textit{et al.} \cite{Lu2021}, but nevertheless remains a consideration. Furthermore, training transformer architectures is computationally expensive, requiring significant computing resources, especially if applied to 3D volumetric medical imaging. In fact, a relatively high number of studies in the field are focused on reducing this computational burden \cite{Xia2023}. This heightened computational resources required thus present a barrier, prohibiting widespread development and applications to new data. Early studies have already demonstrated promising results, and transformer architectures could present a future avenue for spatiotemporal shape modeling. 		
		
\section{Discussion and Conclusions}

	Several approaches for spatiotemporal shape modeling of anatomical structures were discussed in this review. Rapid developments in the field, especially in recent years, have been fueled by advancements in DL and are set to only continually progress further. Nevertheless, the works found in the existing literature have been mainly focused on incremental developments in methodology or applications of novel new tools. This is in contrast with applying already developed tools to existing or novel clinical challenges. This seeming reluctance of the medical imaging community towards application-based research could stem from a multitude of reasons, but a simple lack of data could be the main factor, as we will discuss shortly. Deficiencies notwithstanding, in this section, we will discuss key concepts of spatiotemporal shape modeling uncovered from our review. We will then outline several key barriers to further research in the field before speculating on future research directions. 
	
	\subsection{Nonlinear shape manifolds} 

		From our review, it is clear that anatomical shape variation is highly nonlinear. This nonlinearity is further compounded by the additional nonlinear dynamics of growth and changing biological structures over time, leading to an intricate and complex outlook. Thus, the best-suited models for spatiotemporal progression are those that lie on non-Euclidean manifolds as they best capture this inherently high dimensional problem. A potential reason for this could be the manifold hypothesis, wherein it is postulated that all high-dimensional data lie on an embedded low-dimensional manifold \cite{Fefferman2016, NIPS2010_8a1e808b}. The task of spatiotemporal shape modeling can then be reduced to identifying and characterizing these manifolds, either implicitly or explicitly. The LDDMM framework discussed in Section \ref{sec:LDDMM}, for example, explicitly seeks to uncover spatiotemporal trajectories of diffeomorphisms traversing across a manifold. DL techniques discussed in Section \ref{sec:DL} also implicitly benefit from manifolds, as the efficacy of DL techniques has been attributed to their capability to uncover and disentangle underlying the manifolds of complex data \cite{Brahma2016}.    

		In contrast to manifold-based techniques, several works do exist that have attempted to extend linear (PCA-based) statistical shape models towards spatiotemporal shape models. These, however, fall short when compared to LDDMM and DL-based solutions as they effectively only serve to compare differences across and interpolate between time points as opposed to true longitudinal forecasting \cite{Hamarneh2004, Kasahara2018, BinteAlam2020, Saito2019}. Due to their reliance on landmarks, these methods do not effectively work if anatomies significantly change over time, as is the case, especially in early development. Furthermore, they are incapable of separating groupwise versus individual developmental trends, nor are they capable of effective data imputation \cite{Adams2023}. Therefore, while these methods might be effective for comparing shape variation across time points, they are not as effective for shape trajectory forecasting as manifold-based methods. 
	
		Comparatively, manifold-based techniques are more effective as the longitudinal trajectories traversing the shape space yield an effective description of shape variation over time. LDDMM techniques offer a structured framework to describe shape variation, and furthermore, the geodesic trajectories themselves are clinically relevant as they offer prognostic and diagnostic utility. When utilized within a hierarchical model that incorporates many trajectories for a population average, new trajectories can be estimated for unseen data, which could offer prognostic significance. Furthermore, trajectories can be compared using relational transport operators to diagnose if a trajectory is irregular compared to population averages. Similarly, DL methods mostly operate directly on medical imaging data with convolutional networks. This allows us a way to extract hidden features from images which could also influence spatiotemporal trajectories, otherwise lost during parameterization processes necessary for LDDMM or PCA-based models. In encoding networks especially, the latent space encompassed by these extracted latent variables can be structured to construct a physically meaningful underlying spatiotemporal manifold. The inductive capacity of DL methods with such structured latent spaces is then superior to linear methods, capable of imputing missing data and predicting spatiotemporal trajectories. 
			
	\subsection{Paucity of longitudinal datasets}	
			
		Another clear deficiency is the lack of large, open-source, and high-quality longitudinal imaging datasets. Existing datasets used in studies are generally small, in-house, cover a short time span, and are limited to very specific clinical conditions (Table \ref{tab:my-table}). This is, of course, understandable as it is extremely difficult to gather longitudinal data. Issues such as participant attrition \cite{Young2006} and ethical concerns \cite{Tinker2009} are just two examples of difficulties that hamper the execution of effective studies. An exception to this is the Alzheimer's Disease Neuroimaging Initiative (ADNI) database, which is a large multimodal database of longitudinal biomarker and neuroimaging data tracking the progression of AD \cite{Jack2008}. This dataset is particularly outstanding due to its size and comprehensiveness, leading to many studies covered in this review validating their methods on the ADNI dataset. Nevertheless, this dataset remains unique and standout compared to others. This paucity of longitudinal datasets, especially for medical imaging, impairs the efficacy of both LDDMM and DL techniques covered in this study. 
		
\begin{table}[]
\resizebox{\textwidth}{!}{
\begin{tabular}{lllccll}
\hline
\rowcolor[HTML]{EFEFEF} 
\multicolumn{1}{|c|}{\cellcolor[HTML]{EFEFEF}} &
  \multicolumn{1}{c|}{\cellcolor[HTML]{EFEFEF}} &
  \multicolumn{1}{c|}{\cellcolor[HTML]{EFEFEF}} &
  \multicolumn{2}{c|}{\cellcolor[HTML]{EFEFEF}\textbf{Age range}} & 
  \multicolumn{1}{c|}{\cellcolor[HTML]{EFEFEF}} &
  \multicolumn{1}{c|}{\cellcolor[HTML]{EFEFEF}} \\ \cline{4-5}
\rowcolor[HTML]{EFEFEF} 
\multicolumn{1}{|c|}{\multirow{-2}{*}{\cellcolor[HTML]{EFEFEF}\textbf{Name}}} &
  \multicolumn{1}{c|}{\multirow{-2}{*}{\cellcolor[HTML]{EFEFEF}\textbf{Anatomy}}} &
  \multicolumn{1}{c|}{\multirow{-2}{*}{\cellcolor[HTML]{EFEFEF}\textbf{Modality}}} &
  \multicolumn{1}{l|}{\cellcolor[HTML]{EFEFEF}\textbf{Youngest}} &
  \multicolumn{1}{l|}{\cellcolor[HTML]{EFEFEF}\textbf{Oldest}} &
  \multicolumn{1}{c|}{\multirow{-2}{*}{\cellcolor[HTML]{EFEFEF}\textbf{Number of subjects}}} &
  \multicolumn{1}{c|}{\multirow{-2}{*}{\cellcolor[HTML]{EFEFEF}\textbf{Repeated measurements}}} \\ \hline
OASIS-2                      & Brain      & MRI & 60                   & 96                   & 150                  & \textless{}=5                   \\ \hline
OASIS-3                      & Brain      & MRI & 42                   & 95                   & 1378                 & \textless{}=7                   \\ \hline
HABS-HD                      & Brain      & MRI & 40                   & 92                   & 3838                 & \textless{}=3                   \\ \hline
Harvard Aging Brain Study    & Brain      & MRI & 62                   & 90                   & \textgreater{}290    & \textless{}=3                   \\ \hline
ADNI-4                       & Brain      & MRI & 55                   & 90                   & \textgreater{}2400   & \textless{}=6                   \\ \hline
                             & Brain      & MRI &                      &                      & \textgreater{}5000   &                                 \\
                             & Whole body & DXA &                      &                      & \textgreater{}=5156  &                                 \\
                             & Abdomen    & MRI &                      &                      & \textgreater{}=11365 &                                 \\ 
\multirow{-4}{*}{UK Biobank} & Heart      & MRI & \multirow{-4}{*}{44} & \multirow{-4}{*}{87} & \textgreater{}=5100  & \multirow{-4}{*}{\textless{}=2} \\ \hline
MCSA                         & Brain      & MRI & 30                   & 89                   & 1802                 & ??? \\ \hline                           
\end{tabular}}
\vspace{0.2cm}
\caption{A summary of several longitudinal medical imaging datasets. OASIS - Open Access Series of Imaging Studies. HABS-HD - Health and Aging Brain Study - Health Disparities. ADNI - Alzheimer's Disease Neuroimaging Initiative. MCSA - Mayo Clinic Study of Aging.}
\label{tab:my-table}
\end{table}

		DL techniques are notoriously data hungry, with larger dataset sizes contributing significantly towards improved efficacy of networks \cite{Sun_2017_ICCV, JungKye15}. While techniques such as transfer learning \cite{Alzubaidi2020} and data augmentation \cite{Mumuni2022} seek to ameliorate this issue, it remains pervasive. Conversely, whilst the LDDMM framework is comparatively not as data-hungry, sufficiently sized datasets are also essential. Adequately sized and diverse datasets are vital to ensure that the estimated population average trajectories are reflective of the entire population. Solutions such as GAN-based frameworks discussed in Section 3.2 are shown to be helpful in addressing the issue of data paucity. Therein, generative processes and adversarial training frameworks can increase the generalizability of networks. The latter is particularly useful as the adversarial process assists in regularizing and structuring the latent space, implicitly learning the underlying spatiotemporal manifold. Nonetheless, the lack of datasets presents another issue of validity. In essence, the impressive performance on specific datasets could be a function of the dataset and not the frameworks themselves. Thus, exploring their efficacies on additional anatomical structures and imaging modalities is also prudent. Initiatives to compile multimodal datasets to train and test frameworks in a challenge-like style such as the Medical Segmentation Decathlon (MSD) could be warranted to ensure that future developments in methodology are sufficiently valid \cite{Antonelli2022}. 
		  
		Nevertheless, longitudinal datasets, be it open-source or in-house, remain scarce. Gathering additional longitudinal data remains the most ideal option, however the aforementioned practical difficulties in data gathering present a significant barrier. In the medium to long term, additional initiatives resembling ADNI could be warranted to gather high-quality, longitudinal, multi-center data for other diseases and disorders benefiting from spatiotemporal shape analyses. Furthermore, these data should be multi-modal, encompassing both imaging and also biomarker data as these have been shown to work synergistically when incorporated into joint frameworks, improving their efficacy. In the meantime, efforts to compile existing data into a large open-source database could be more warranted. This could resemble, for example, the aforementioned MSD. Nonetheless, the impetus to gather and unite such datasets is lacking, especially in the face of general (un)willingness to openly share rare datasets \cite{Tedersoo2021}.

	\subsection{Future outlook and directions}

		In this review, we focused mainly on the development of LDDMM and DL-based techniques. We did so because these were considered the most versatile for generalizable spatiotemporal shape modeling. This is opposed to alternative methods seeking to model shape changes of specific anatomical structures over time from a mechanistic standpoint. For example, many early works on spatiotemporal shape modeling of tumors attempted to develop models uncovering the underlying mechanistic cause and effects governing their growth \cite{Jarrett2018}. Similar works also exist focusing on cardiac tissue remodeling \cite{Wang2017333333} and bone remodeling \cite{Kameo2020}. Whilst these varied mechanistic models are inherently different, they generally revolve around shape change as a consequence of mechanical and biochemical stimuli or a combination thereof. Thus, these models seek to uncover the underlying formulae governing these interactions and their relationships. This is in contrast with the LDDMM framework, which operates solely from a geometric perspective in uncovering the trajectories of diffeomorphic transformations. In other words, the LDDMM framework does not explicitly consider the underlying physical laws governing the biological processes that lead to the resulting shape changes. Therefore, this approach potentially neglects key information that may affect how reflective the LDDMM approach is in said processes and, therefore, its accuracy. Similarly, DL techniques are opaque, often referred to as black boxes \cite{Castelvecchi2016}. Therein, the model layers, in effect, operate on hidden features uncovered during training processes. These have, in principle, no physical meaning and are not always explainable, engendering issues of trust and validity. A compromise and potential future direction of research is via physics-informed neural networks (PINNs) \cite{Cuomo2022}. Therein, the strengths of DL to process large datasets are utilized to solve underlying physical equations that describe the physics of a system. PINNs are particularly useful even, for example, to uncover underlying dynamics of systems that were previously obscured under high dimensional nonlinear data \cite{Lagergren2020}. Tajdari \textit{et al.} demonstrated the applicability of PINN principles within frameworks to model the longitudinal progression of adolescent idiopathic scoliosis, outperforming traditional methods \cite{Tajdari2021, Tajdari2022}. While their works were mainly concerned with the mechanistic effects of loading on spinal outcomes, their efficacy also lends itself to potential benefits for spatiotemporal shape modeling. Nevertheless, PINNs remain an unexplored avenue and warrant further study. 
		
		Another developing field is utilizing and exploiting causality in the form of causal deep learning. In essence, causality and structural causal models (SCMs) seek to capture and model the chain of causality and inter-variability of multivariate systems \cite{peters2017elements, Pearl2013, pearl2016causal}. This is useful as it enables us to interrogate models to obtain counterfactuals (\textit{i.e.,} if $X$ was different, what is the effect on $Y$? Or more relevantly, 'How would injury $A$ affect bone development of pediatric subject $B$?'). For spatiotemporal shape modeling specifically, this could be used to obtain predictions of shape change over time as a counterfactual from existing data. Traditional methods relied on a system of structural equations with computation graphs, but this precluded the use of higher dimensional data such as images. In recent years, several studies have explored extending SCMs towards being supported by DL (\textit{i.e.,} Deep Structural Causal Models (DSCMs)), enabling the use of hidden features identified via DL \cite{Lore2018, kjsadflkjaslkdjfsdf, heeeeeeeeeeeeeeeeeeeee}. Zhou \textit{et al.} reviewed the synergistic capabilities of generative models and causality, specifically highlighting the applicability of the latter in enhancing the interpretability of generative processes \cite{hehehehehehehhehehehe}. Further works such as by Reinhold \textit{et al.} demonstrated the capability of DSCMs to generate counterfactual brain MRIs of patients with multiple sclerosis \cite{Reinhold2021}. They were able to manipulate demographic and disease covariates and observed their effects on MRI imaging in a novel proof-of-concept. Rasal \textit{et al.} further extended DSCMs towards shape modeling, specifically 3D meshes, demonstrating the extendability and scalability of the principle towards more complex data types \cite{Rasal2022}. Nevertheless, the field is still in its relative infancy, with further developments and refinements in the DSCM framework potentially leading to enhanced efficacy for longitudinal shape modeling. 

        Finally, another potential avenue for development is \textit{via} diffusion models (DMs) \cite{FuestMaGui24, 10081412, Kazerouni2023}. Similar to GANs and AEs described earlier, DMs are a form of generative model where a forward and inverse process are learned to reconstruct input data. In contrast with the aforementioned network architectures, DMs instead function on the principle of noise addition and removal. DMs consist of forward and reverse processes, wherein noise is added onto input data in successive steps and the resulting noise is reversed to reform the input data \cite{DDPM20}. DMs have led to state-of-the-art high resolution visual generative networks \cite{HRIS2021, DMBGIS21}, and existing works have demonstrated the potential for use in a variety of spatiotemporal modeling tasks \cite{ASDMTSSTD2024, NEURIPS2023_8df90a14}. The use of DMs for spatiotemporal shape modeling is still relatively unexplored, with most studies focused on alternative spatiotemporal data (\textit{e.g.,} weather forecasting, traffic patterns, ECG). Nonetheless, Yoon \textit{et. al.} demonstrated the use of a sequence aware diffusion model (SADM) to generate longitudinal medical images \cite{Yoon2023}. Their framework utilized a sequence aware transformer as the conditional module for a diffusion model, demonstrating effective data generation capabilities for longitudinal 3D medical imaging sequences, even with missing data.

		In summary, this paper mainly reviewed the LDDMM framework and DL-based techniques for longitudinal shape modeling. Both achieve their remarkable state-of-the-art performance as they function on similar principles of uncovering the underlying nonlinear spatiotemporal data manifold. Whilst promising, the LDDMM framework is computationally expensive and inefficient due to the exhaustive optimization procedure necessary to calculate smooth and invertible diffeomorphisms. It, nevertheless, demonstrates strong capabilities to establish hierarchical models that differentiate individual and population-level temporal trajectories. Conversely, DL-based techniques are powerful but data-hungry and lack underlying physical meaning. Network architectures have been developed to predict shape changes in anatomical structures. Nevertheless, the underlying data manifolds and spatiotemporal trajectories governing these predictions are obscured by the 'black box' nature of DL architectures. This affects the interpretability of these predictions, especially if the longitudinal trajectories themselves are important. Nevertheless, the capability of DL architectures to identify hidden features from input images and implicitly map the underlying data manifolds denote their importance for spatiotemporal shape modeling. Our review highlights that hybrid techniques that amalgamate both approaches' strengths are more desirable. Furthermore, frameworks incorporating multi-modal data improved generalizability. Thus, further works should not neglect the utility of auxiliary data (\textit{e.g.,} biomarker levels, demographic information, \textit{etc.}). Finally, we theorize that utilizing mechanistic models in a manner similar to PINNs or structured causal frameworks could also further improve the predictive capacities of future spatiotemporal shape models.

\bibliographystyle{IEEEtran}
\bibliography{references.bib}

\end{document}